\documentclass{tlp}

\usepackage{amsmath}
\usepackage{graphicx}
\usepackage{multirow}

\usepackage{url}
\usepackage{xspace}
\usepackage{microtype}
\usepackage{fvextra}
\usepackage{algorithm}
\usepackage{algorithmic}
\usepackage{listings}
\usepackage{booktabs} 
\usepackage{hyperref}
\usepackage{subcaption}
\usepackage{multirow}
\usepackage{makecell}
\usepackage{fancyvrb}
\definecolor{keywordcolor}{rgb}{0.0, 0.0, 1.0} 
\definecolor{commentcolor}{rgb}{0.0, 0.5, 0.0} 
\definecolor{stringcolor}{rgb}{1.0, 0.0, 0.0} 

\newcommand{\iec}{i.e.,\xspace}
\newcommand{\egc}{e.g.,\xspace}
\newcommand{\graph}{\ensuremath{\mbox{CLEGR}^V}\xspace}
\newcommand{\asplarrow}{\mathrel{\mathrm{:\!\!{-}}}}

\begin{document}

\lefttitle{Thomas Eiter, Nelson Higuera, Johannes Oetsch}

\jnlPage{1}{?}
\jnlDoiYr{2025}
\doival{10.1017/xxxxx}



\title[Distilling ASP Rules for Neurosymbolic VQA]{Distilling Answer-Set Programming Rules from LLMs for Neurosymbolic Visual Question Answering}

\begin{authgrp}
  \author{\sn{Eiter} \gn{Thomas}}
  \affiliation{Vienna University of Technology \textup{(}TU Wien\textup{)}, Favoritenstrasse 9--11, Vienna, 1040, Austria \\
  \email{thomas.eiter@tuwien.ac.at}}
  
  \author{\sn{Higuera Ruiz} \gn{Nelson}}
  \affiliation{Vienna University of Technology \textup{(}TU Wien\textup{)}, Favoritenstrasse 9--11, Vienna, 1040, Austria \\
  \email{nelson.ruiz@tuwien.ac.at}}
  
  \author{\sn{Oetsch} \gn{Johannes}}
  \affiliation{Jönköping University, Gjuterigatan 5, 551\,11 Jönköping, Sweden \\
  \email{johannes.oetsch@ju.se}}
\end{authgrp}

\maketitle
 
\begin{abstract}
Visual Question Answering (VQA) is the task of answering questions about images, requiring the integration of multimodal input and reasoning. Modular approaches that incorporate logic-based representations into the reasoning component offer clear advantages over end-to-end trained systems, particularly in terms of interpretability. However, adapting or extending these representations when task requirements change can place a significant burden on developers. To address this challenge, we present an approach for distilling rules from Large Language Models (LLMs). Our method prompts an LLM to extend an initial VQA reasoning theory, expressed as an answer-set program, to meet new requirements of the task. Examples from VQA datasets guide the LLM, validate the results, and help correct erroneous rules by leveraging feedback from the ASP solver. We demonstrate that our approach is effective across diverse VQA datasets. Notably, only a few examples are needed to elicit correct rules from LLMs. Our experiments suggest that rule distillation from LLMs is a promising alternative to traditional data-driven rule learning approaches.
Under consideration in Theory and Practice of Logic Programming (TPLP).
\footnote{The code for reproducing our experiments is available as an online repository: \url{https://github.com/pudumagico/KDASP}.} 
\end{abstract}



\section{Introduction}\label{sec:intro}

\emph{Visual Question Answering} (VQA)~\citep{antol2015vqa,goyal2017vqav2} is a challenging problem with important applications~\citep{barra2021vqaapps,DBLP:journals/artmed/LinZTSHWHG23, DBLP:journals/access/DingYLS25}; it 
requests to provide an accurate answer for a question about a visual scene. This requires not just a joint understanding of vision and text, but also  
to follow complex chains of reasoning operations. 

\emph{Neurosymbolic approaches to VQA}~\citep[etc.]{MaoGKTW19,yi2019neuralsymbolicvqa,amizadeh2020neurosymbolic,eiter2022neurosymbolicasp,vipergpt,JohnstonNS23a} use deep learning for perception, produce a symbolic representation of the input image and question, and then perform reasoning on this representation in a purely symbolic way. 
These approaches are interpretable, transparent, and can be extended due to their compositional structure. 
%
A promising direction in this regard is to use logic-based formalisms for the reasoning component. 
We are in particular interested in using Answer-Set Programming (ASP)~\citep{brewka2011asp,lifschitz2019answer}, a prominent knowledge representation formalism, to realise the reasoning module of such systems. In addition to concise representations, advantages are that 
non-determinism allows for multiple answers if desired,
ambiguity in perception often can be resolved by reasoning~\citep{eiter2022neurosymbolicasp}, and one can more easily add explanation capabilities~\citep{EiterGHO23,eite-etal-xvqa-25}.  
ASP for augmenting VQA with reasoning capabilities is in fact gaining traction~\citep{AbrahamAR24,EiterGHO23,eiter2022neurosymbolicasp,BasuSG20,e64e46e012e944c399cb4a0b2af7ca58}, while outside of VQA it is used for visual tasks such as segmentation of medical images~\citep{brunoCMM21} and compliance checking for control panels~\citep{barbara2023neuro}; see the recent survey by \cite{10.1007/978-3-031-89366-7_1}.


The downside is that maintaining or extending ASP representations when requirements
change, \egc when there is a change in the types of question that needs to be answered, can be an additional burden on the developer.
We address this issue by presenting an approach for \emph{declarative knowledge distillation} from Large Language Models (LLMs)~\citep{NIPS2017_3f5ee243,DBLP:journals/corr/abs-2303-18223}.
The premise of this work is that we have a neurosymbolic VQA system in place whose reasoning component involves an initial ASP theory. 
If there is a shift in requirements, reflected by new examples for which the system does not yet give the correct answer, our method is to prompt an LLM to extend this initial theory to meet the new requirement.
Hence, examples from a VQA dataset are used to guide the LLM in generating new rules and validate the results, which involves using multiple prompting strategies, mending the generated rules if they are not correct using feedback from the ASP solver, and regressively testing past examples to ensure consistency.

Statistical-relational learning (SRL)~\citep{RaedtK17} aims to learn rules that generalize from dataset examples. A notable subfield of SRL is Inductive Logic Programming (ILP)~\citep{DBLP:journals/ml/CropperDEM22}, where systems rely on a hard search space—explicitly defined by the developer through mode declarations or language bias. In contrast, LLMs operate within a soft search space implicitly encoded via natural language prompts. 
Our method leverages this capability by using examples not as training data, but solely to guide a knowledge distillation process that conditions the LLM toward suitable rule completions (see Section~\ref{sec:rel} for further discussion). 
Notably, our experiments show that only a few examples are sufficient to elicit correct rules from the LLM. This distinguishes our approach from data-driven methods, which typically require logical elaboration or substantially more training data.


We use a diverse selection of VQA datasets to evaluate our knowledge distillation method: GQA~\citep{hudson2019gqa}, CLEVR~\citep{johnson2017clevr}, and \graph~\citep{
bauer2023, bauer24}.  GQA uses real images that depict complex visual scenes and questions with a large number of possible answers that involve processing objects, their attributes, and relations between them. CLEVR uses synthetic scenes of geometric objects and is designed to challenge VQA systems with compositional questions that require breaking down the overall task into 
an evaluation graph of primitive operations. \graph is a recently published dataset that uses images of graphs that resemble transit networks and requires graph-based reasoning.
To test our approach, we adopt the representations from modular VQA systems with ASP for reasoning that can solve these datasets~\citep{EiterGHO23,Hadl23,bauer2023,bauer24}.


Our main contributions are briefly summarized as follows:
\begin{itemize}
    \item[-] We propose a method for \textit{declarative knowledge distillation} from LLMs, enabling the automated extension of ASP-based reasoning modules in VQA systems, including the ability to compose new ASP rules from existing ones.
    \item[-] We design prompting strategies for the distillation method that includes multi-prompting, chain-of-thought, and rule mending.
    \item[-] We experimentally evaluate our method across three diverse VQA datasets---GQA, CLEVR, and \graph. Our evaluation includes also ablation studies involving the proposed prompting strategies.
    \item[-] We introduce a redundancy elimination heuristic that attempts to prune the distilled rule sets without reducing accuracy.
\end{itemize}

Our experiments show that LLMs can grasp and produce ASP rules to implement new reasoning operators, and thus can effectively assist in the process of extending a reasoning component. 
Ablation studies reveal that the effectiveness of prompting strategies depends on both the model and the dataset: while stronger models benefit significantly from their use, smaller models may be hindered by certain strategies.
This methodology offers a promising avenue for extracting domain-specific knowledge from LLMs in the realm of VQA and offers a viable complement to data-driven rule learning approaches.


\section{Background}\label{sec:background}

The logic-based VQA approaches for the VQA datasets that we are going to use for our evaluation use a modular architecture and ASP to derive answers from a symbolic scene and question representation. 
Before we provide background on these VQA approaches, we review the basics of ASP.

\subsection{Answer-Set Programming}
    
Answer-Set Programming (ASP)~\citep{brewka2011asp,lifschitz2019answer}
is a well-known approach to declarative problem solving, in which 
solutions to a problem are described by programs that consist of sets of logical rules. Efficient ASP solvers for evaluating programs are available such as clingo \cite{gebser2019multi}, dlv \cite{leone2006dlv}, or wasp \cite{DBLP:conf/lpnmr/AlvianoDFLR13}.

For our concerns, an ASP program is a finite set $P$ of rules $r$ of the form
\begin{equation}\label{asp-rule}
a \asplarrow\ b_1,
\ldots,\ b_n,\ not\ c_1, \ldots,\ \mathit{not}\ c_n\nonumber \quad m,n \geq 0,\,
\end{equation}
\noindent where $a$ (the \texttt{head}), all $b_i$, and all $c_j$ (the \texttt{body}) are atoms in a
first-order predicate language, and $\mathit{not}$ stands for negation
as failure (weak negation). We allow that $a$ may be missing (viewed as falsity);
then $r$ acts as a constraint. 
Intuitively, the rule means that whenever all $b_i$ are true and none
of the $c_j$ can be shown to be true, then $a$ must be true.  
A \emph{fact} is a rule with empty body (in which case $\asplarrow$ is omitted). 
Facts represent knowledge that is unconditionally true.

The semantics of a ground (variable-free) ASP program is given
by answer sets, which are Herbrand models that satisfy a stability
condition~\citep{GelfondL88}.  A Herbrand interpretation of $P$ is a set $I$ of ground
atoms in the language 
of $P$ (intuitively, the atoms that are
true); 
it is a model of $P$ if for each rule $r\,{\in}\,P$ either
\begin{enumerate}[label = (\roman*)]
\item $a \in I$ or 
\item  $\{b_1,\ldots,b_n\}\not\subseteq I$ or
\item  $I\cap \{c_1,\ldots,c_n\}\neq \emptyset$; \iec $I$ satisfies $r$
viewed as implication in classical logic. 
\end{enumerate}
%
Furthermore, $I$ is an answer set of $P$ \citep{DBLP:conf/jelia/FaberLP04}%
\footnote{This definition is equivalent to the one of \cite{GelfondL88} for programs consisting of rules (\ref{asp-rule}) and ensures benign semantics of recursive aggregates \citep{DBLP:conf/jelia/FaberLP04}.} 
if  $I$ is a $\subseteq$-minimal model of the program 
\[ 
P^I = \{ r \in P \mid I \mbox{ satisfies neither } (ii) \mbox{ nor } (iii) \}\,. 
\]
Intuitively, 
$I$ must result by applying the rules $r$ whose bodies
``fire'' w.r.t.\ $I$ starting from facts.
The semantics of programs with variables is defined in terms of their
groundings (uniform replacement of variables in rules with all
possible ground terms).

ASP provides further constructs that extend its modeling capabilities:
\emph{choice rules} allow for selecting a subset of atoms under cardinality bounds. They are of the form
\[
l \ \{ a_1,\ldots,a_k \} \ u \;\asplarrow\ \texttt{body}
\]
where $l,u$ are integers specifying the minimum and maximum number of atoms $a_i$ 
that may be true if the \texttt{body} holds. This supports controlled nondeterminism and is useful for selection problems.

\emph{Aggregates} enable reasoning over sets of terms, for instance, to impose conditions on their cardinality or sums. Here is an example of a rule that involves an aggregate expression for counting:
\[
p(X) \;\asplarrow\ \#\texttt{count} \{ Y : q(X,Y) \} > N.
\]
This example states that $p(X)$ holds if the number of $Y$ such that $q(X,Y)$ holds exceeds $N$. 
ASP supports various aggregates, including \texttt{\#sum}, \texttt{\#min}, and \texttt{\#max}.

While a hard constraint is a constraint that must be satisfied by any feasible solution to the ASP program, a \emph{weak constraints} (soft constraint) can be violated, but violating it incurs a penalty in the objective function. A weak constraint has the following form:
\[
:\ \!\!\sim \texttt{body}.\; [w@l]
\]
where $w$ is the penalty cost and $l$ the priority level.
ASP solvers seek answer sets that minimize the total penalty incurred by the weak constraints whose body is satisfied. Ties are broken using the priority level.
For more details on ASP, we refer to \citep{brewka2011asp,lifschitz2019answer,calimeri2012asp}.

\subsection{Modular VQA with Answer-Set Programming}


\begin{figure*}[t!]
    \centering \small
    \begin{tabular}{ccc}
    \includegraphics[height=2.7cm]{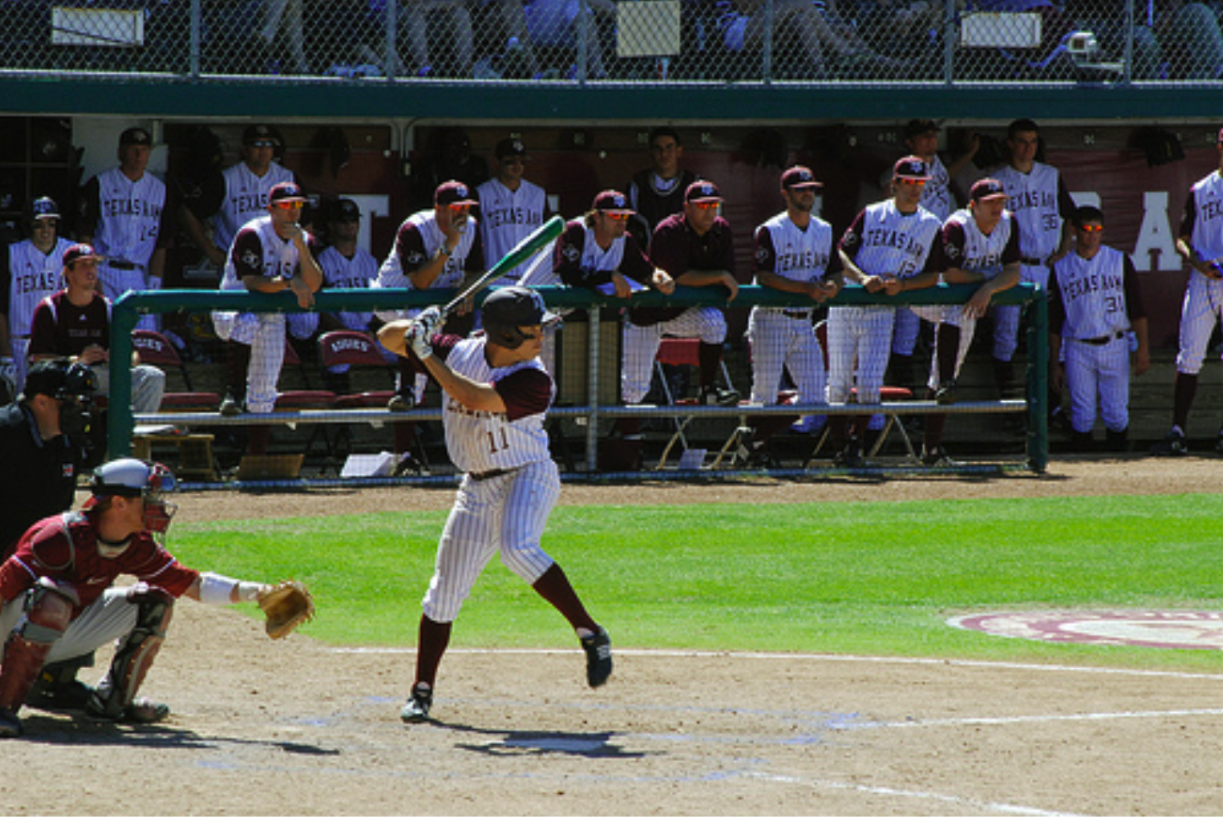} &
    \includegraphics[height=2.7cm]{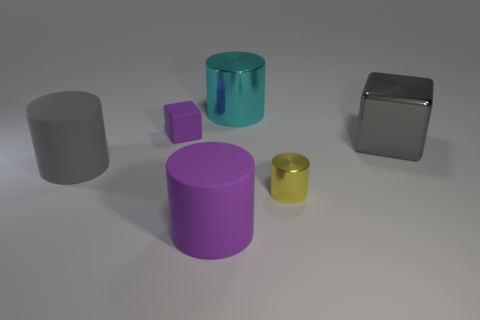} &
    \includegraphics[height=3.5cm]{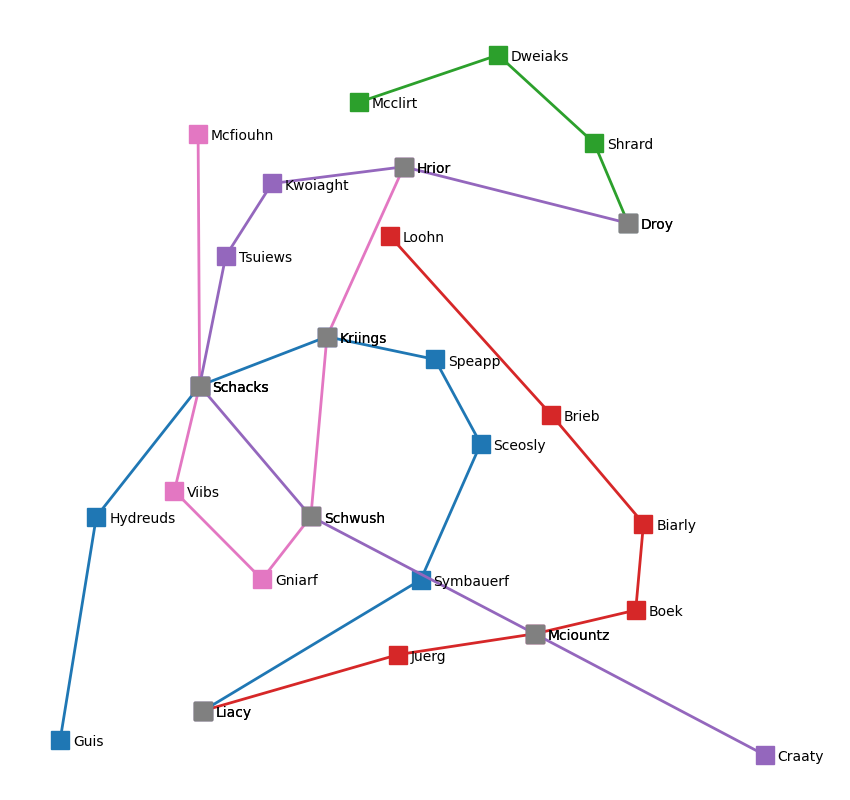} \\
    \begin{minipage}{0.3\linewidth}
    Is the umpire to the right or to the left of the standing person that is wearing a helmet?  
    \\
    \textbf{Answer:} To the left.
    \end{minipage} &
    \begin{minipage}{0.3\linewidth}
    Is there an object of the same color as the small cube?
    \\
    \textbf{Answer:} Yes.
    \end{minipage} &
    \begin{minipage}{0.3\linewidth}
    How many stations are between Loohn and Boek?
    \\
    \textbf{Answer:} 2.
    \end{minipage} \\
    (a) & (b) & (c)
    \end{tabular}

    \caption{Example scenes and questions from GQA (a), CLEVR (b), and \graph (c). }
    \label{fig:vqa-examples}
\end{figure*}

We consider three VQA datasets as application area for our rule distillation approach: GQA~\citep{hudson2019gqa}, CLEVR~\citep{johnson2017clevr}, and \graph~\citep{bauer2023,bauer24}. 
Example instances of these datasets are shown in Fig.~\ref{fig:vqa-examples}. 
They involve realistic scenes in GQA, synthetic scenes in CLEVR, and images of labeled graphs in \graph.

\begin{figure}
    \centering
    \includegraphics[width=1\linewidth]{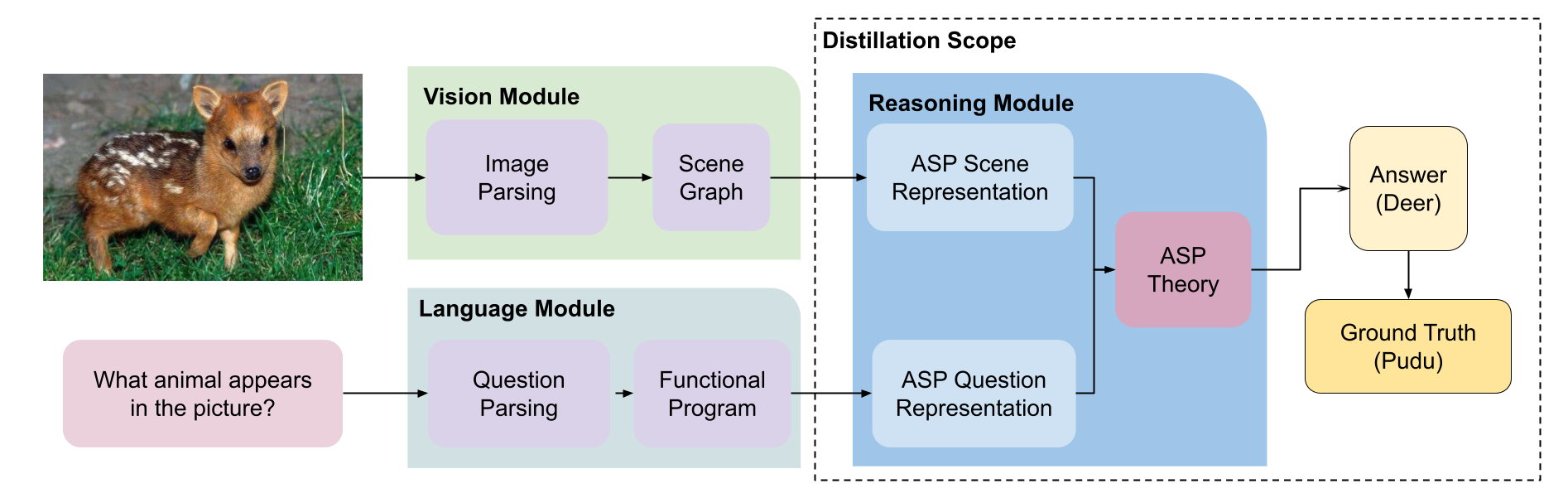}
    \caption{Generic Neurosymbolic VQA pipeline with three main components: (i) vision module, (ii) language module and (iii) reasoning module. Our distillation method operates on the symbolic representations of scenes and questions, ASP theory, as well as the ground-truth answers to the questions.}
    \label{fig:generic_nsvqa}
\end{figure}

For our distillation process, we adopt the logical representations and encodings of neurosymbolic systems that use ASP for reasoning, and instantiate a generic modular neurosymbolic architecture, shown in 
Fig.~\ref{fig:generic_nsvqa}
that has the following core components:

\begin{itemize}
\item {\bf Vision module:} Neural networks that process the visual scene to produce a symbolic representation of the scene, called the \emph{scene graph}. It contains a structured representation of each object in the image, with their corresponding attributes and relationships.

\item {\bf Language module:} Neural networks that process the natural language question to produce a symbolic representation of the question, called a \emph{functional program}. This is a tree-like structure, where every node defines a \emph{primitive function}, and edges between the nodes represent the flow of information from the leaves to the root.

\item {\bf Reasoning module:} A component that uses an ASP solver to derive the answer. 
First the scene graph and functional program are translated into sets of ASP facts.
Details of these representations depend on the dataset, but
they have in common that each primitive function of the functional program is represented by a single ASP fact in the question representation. 
We use an ASP program, which we also refer to as ASP theory, to encode how to derive an answer based on scene and question encodings.
An ASP solver computes then an answer set from which the answer to the question is  extracted.
\end{itemize}

The ASP theory is a set of rules that---in contrast to the question and scene encoding---does not change from question to question. The theory is thus {\em general}\/ and encodes the semantics of the primitive functions in a \emph{uniform}\/ way. 
Using ASP as the symbolic formalism for reasoning provides us not only with a mature ecosystem of tooling and solvers
but, and notably, allows one to capture the uncertainty in the predictions of the image-processing component. 

The collection of primitive functions the ASP theory implements depends on the task. 
For GQA, the focus is on reasoning on a complex scene graph, CLEVR focuses on processing compositional questions, and \graph involves multi-modal reasoning tasks on labeled graphs, \egc deciding reachability between nodes with certain labels.
Also, the details of how the vision and language module are realized varies from one application to another. We next review some of these details and the characteristics of the datasets used. For each dataset, we present examples of a scene and question as ASP facts, and the complete ASP theory in~\ref{apdx:theories}.

\subsubsection{The GQA dataset}

We use the state-of-the-art GQA dataset, which has been widely adopted
recently~\citep{amizadeh2020neurosymbolic,vipergpt,liang2020lrta,li2023blip2}. 
It contains over 22M open and binary questions that are complex in structure, involve a wide variety of reasoning skills, and have a large number ($1\,878$) of possible answers.  

The questions cover more than $100\,000$ images from the Visual Genome dataset \citep{krishna2016visualgenome} that present real-world scenes with a wide variety of object classes, attributes, and relations. 
GQA provides both a functional representation of each natural-language question and a Visual Genome scene graph for every image, which allows us to construct ASP programs that always infer the correct answer whenever there is perfect language and visual information. We later will erase parts of these programs to simulate incomplete correct programs and ask an LLM to extend them accordingly.

We adopt the representations and theory of a
recent system for GQA~\citep{Hadl23,eite-etal-xvqa-25} that 
adheres to the general architecture described above. 

For the vision module, the system uses {\em ``question-driven'' partial scene-graph extraction}, where only information is extracted from the scene that is relevant for answering the question at hand. 
It does so by relying on the foundation models CLIP~\citep{radford2021clip} and OWL-ViT~\citep{minderer2022owlvit} for processing the visual scene. They do not need any training and thus solve the task in a zero-shot manner.
A snippet of ASP facts from an ASP scene representation is: 

{
\begin{Verbatim}[breaklines=true, breakanywhere=true]
object(o1).
has_attr(o1, class, alive). 
has_attr(o1, class, person). 
has_attr(o1, class, baseball_player). 
has_attr(o1, name,  baseball_player).
\end{Verbatim}
}

These facts describe parts of a scene that contains an object whose most specific class is ``baseball\_player'' (cf.~Fig.\ref{fig:vqa-examples}).

Questions are translated from their functional program into ASP facts.
For example, the question ``Is the umpire to the right or to the left of the
standing person that is wearing a helmet?''
may be rendered as follows:

{ 
\begin{Verbatim}[breaklines=true, breakanywhere=true]
scene(0).
select(1, 0, helmet).
relate(2, 1, person, wearing, subject).
filter_any(3, 2, standing).
choose_rel(4, 3, umpire, to_the_left_of, to_the_right_of, subject).
\end{Verbatim}
}

The numbers encode steps of an evaluation plan for deriving the answer, where 
the first argument is the current step,
followed by numerical input steps providing input and/or additional arguments. All of our ASP question encodings adopt variations of this representation.
In comparison with the original functional program, each of these facts represents a primitive function, while numbers, in them represent states, denoting the flow of computation from state to state, serving virtually as edges in a graph where the nodes are the primitives.

The ASP theory encodes the semantics of the
primitive functions that can appear 
in the question
encoding. 
The whole theory has {60} rules;
some rules are:

{
\begin{Verbatim}[breaklines=true, breakanywhere=true]
state(TO,ID)    :-  scene(TO), object(ID).
state(TO,ID)    :-  select(TO,TI,C), state(TI,ID), has_attr(ID,_, C).
state(TO,ID)    :-  filter_any(TO, TI, VALUE), state(TI, ID), 
                    has_attr(ID, ATTR, VALUE).
\end{Verbatim}
}
In these rules,  \verb!TI! and \verb!TO! are variables representing input/output step references, \verb!ID! represents an object id, and \verb!C! an object class. 
These rules define how the state of candidate objects is updated based on the scene, selection criteria, and attribute filters. 
For example, the \texttt{select} rule propagates objects from a previous step \verb!TI! to the current step \verb!TO! if they have the specified class \verb!C!.

The entire theory is solved alongside both scene and question encodings to produce an answer. 

\subsubsection{The CLEVR dataset} 
The second dataset we consider is CLEVR~\citep{johnson2017clevr}, which uses synthetic scenes but challenges VQA systems with long compositional questions. 
The ASP-based VQA system used for CLEVR has been extended by an explanation component~\citep{eiter2022neurosymbolicasp,EiterGHO23}. 
The dataset contains synthetically generated images with different objects.  
The latter vary in their shape, 
color, 
size, and 
material. 
CLEVR questions require, \egc identifying objects, counting, filtering for attributes, comparing attributes, and spatial reasoning. 
They are formulated in natural language, but, as for GQA, a functional representation is also provided that can be directly parsed into ASP facts.

The architecture of the system for CLEVR from where we take the ASP theory is similar to the one for solving GQA.  
However, the scenes are less complex as geometric objects are clearly distinguishable, using the
popular object-detection framework YOLOv5\footnote{\url{ultralytics.com/yolov5}.} to
identify all objects in the image, their attributes, and bounding-box information. 




A snippet of ASP facts from a scene representation, question representation, and theory are: 

{
\begin{Verbatim}[breaklines=true, breakanywhere=true]
obj(0,129,121,rubber,green,cube,small).
obj(1,131,202,metal,blue,cylinder,small).
obj(2,247,89,rubber,green,cube,small).
\end{Verbatim}
}

These facts describe objects and their different attributes, such as ID, center of the bounding box $X,Y$ positions, material, etc.
The question ``Is there a small green cube in the scene?''
is rendered as follows:

{ 
\begin{Verbatim}[breaklines=true, breakanywhere=true]
scene(0).
filter_small(1).
filter_green(2).
filter_cube(3).
unique(4).
exists(5).
end(6).
\end{Verbatim}
}

In a similar fashion to GQA, the numbers encode evaluation steps.
The ASP theory consists of {72} rules;
some of the rules are:

{
\begin{Verbatim}[breaklines=true, breakanywhere=true]
object(ID) :- obj(ID,_,_,_,_,_,_).
has_size(ID,SIZE) :- obj(ID,_,_,_,_,_,SIZE).
has_color(ID,COLOR) :- obj(ID,_,_,_,COLOR,_,_).

state(T+1,ID) :- scene(T), object(ID).
state(T+1,ID) :- filter_small(T), state(T,ID), has_size(ID,small).
state(T+1,ID) :- filter_green(T), state(T,ID), has_color(ID,green).
state(T+1,ID) :- unique(T), state(T,ID).
:- unique(T), state(T,ID), state(T,ID'), ID!=ID'.
\end{Verbatim}
}
Here, the \texttt{unique} rule propagates exactly one object from step \texttt{T} to \texttt{T+1}, enforced by a constraint that eliminates any answer set where multiple objects are present at step \texttt{T}.
The rules follow essentially the same pattern as those for GQA, evaluating some predicate in a state and then passing this information to the following state.
\subsubsection{The \graph dataset} 
 
The recently introduced \graph dataset~\citep{bauer2023,bauer24} consists of images of graphs that resemble transit networks;
the task is to answer questions concerning such graphs.  An illustration of a graph and a question that requires to find the shortest path between two stations and count the number of stations that are on that path is shown in Fig.~\ref{fig:vqa-examples}.



We adopt the ASP theory of the modular neurosymbolic architecture from the same publication. 
The vision module is multi-modal in the sense that it deals with graph and text information: it combines \emph{optical graph recognition} (OGR)~\citep{auer2013ogr} for graph parsing,
a tool that parses the graph image into an abstract representation that contains nodes and edges, and
a pretrained \emph{optical character recognition} (OCR) neural network~\citep{abin2018ocr} to parse the image into text labels that are associated with the closest node.
The language module parses the natural language questions using either regular expressions or LLMs.
Finally, the ASP theory encodes the semantics of the question as a logic program.

A snippet of ASP facts from a scene representation, question representation and theory are: 

{
\begin{Verbatim}[breaklines=true, breakanywhere=true]
node(thraonk,brown).
node(mcciourly,orange)
edge(shuosh,krauess,olive).
edge(krauess,sairr,olive).
line(brown).
line(olive).
\end{Verbatim}
}

These facts describe the different nodes, edges and the ``metro lines'' the nodes and edges belong to, \egc the node ``Thraonk'' is in the brown line.
The question ``How many stations are in the shortest path between Thraonk and McCiourly?''
is encoded as follows:

{ 
\begin{Verbatim}[breaklines=true, breakanywhere=true]
station(0,thraonk).
station(0,mcciourly).
shortestPath(1).
countNodesBetween(2).
end(3).
\end{Verbatim}
}

Again, the structure resembles the same evaluation procedure of GQA and CLEVR.
This theory consists of {56} rules;
some of the rules are:

{
\begin{Verbatim}[breaklines=true, breakanywhere=true]
sp((N, N')) :- shortestPath(T), station(T-1,N), station(T-1,N'), N<N'.
0 { selected((N, N'),S1,S2); 
    selected((N, N'),S2,S1) } 1 :- edge(S1,S2,C), sp((N, N')).
path((N, N'),X,Y)   :- selected((N, N'),X,Y).
path((N, N'),X,Z)   :- path((N, N'),X,Y), path((N, N'),Y,Z).
:- sp((N, N')), not path((N, N'), N, N').
cost((N, N'), C)    :- sp((N, N')), 
                       C = #count { X,Y : selected((N, N'),X,Y) }.
#minimize { C,N,N' : cost((N, N'), C) }.
\end{Verbatim}
}


The first rule identifies the station pair \texttt{(N, N')} for which a shortest path is requested, specified by the question encoding using the \texttt{shortestPath} predicate. 
The second rule uses a choice rule to non-deterministically select directed edges that may form a path between the stations in the pair, allowing at most one direction of the edge. 
The next two rules define the transitive closure over the selected edges, determining which stations are reachable from each other via the current selection. 
The constraint ensures that the chosen edges indeed connect the given station pair, eliminating answer sets that fail to do so.
The penultimate rule computes the number \texttt{C} of selected edges used for each station pair using the \texttt{\#count} aggregate.
Finally, the \texttt{\#minimize} statement ensures that among all valid edge selections, the one with the smallest number of edges is preferred.

We note that the initial ASP theory is completely manually constructed. 
In our experiments, we simulate the extension process by removing rules from a ``perfect'' theory, \iec one that always calculates the ground truth answer, to create an incomplete version.
This means that we only use theories we know are correct; using faulty initial theories is not in the scope of this article but is an interesting possible direction that involves the remediation of an ASP program.

\section{Knowledge Distillation Method}\label{sec:distillation}

Our VQA approaches start from an initial ASP theory that is constructed for a particular dataset. 
Given a correct representation of both an image and question, the theory is always able to compute the correct answer. 
If the dataset is extended in a way that the theory no longer can answer correctly, rules need to be modified or added to handle new examples. 
To this end, we propose an LLM-based system to aid and provide automated support.

We aim for a reasoning module that manages the theory and is able to recognise which examples it can handle. If an example cannot be solved, the module updates the theory. We propose a method of \emph{declarative knowledge distillation}, where the model we distill from is an LLM, and the distilled knowledge is represented as ASP rules.
Our method involves different prompting strategies to obtain rules and to correct the incorrect answers of the LLM by using feedback from the ASP solver.

\subsection{Rule Distillation Algorithm}

\begin{algorithm}[t]
\caption{Rule distillation from LLMs.}
\label{alg:ruledistill}
\begin{lstlisting}[language=Python]
# T: initial ASP theory
# distillation_suite: list of examples of form 
#          (scene, question, answer)
regression_examples = []
for ex in distillation_suite:
  s, q, a = ex
  if asp_solver(T + s + q) == a:
    continue

  rules  = llm(multiprmpt, T, ex)
         + llm(CoT_simple, T, ex)
         + llm(CoT_complx, T, ex)

  for r in rules:
    if not syntax_check(T + r, ex):
      r = llm(syntx_prmpt, T + r, ex)
      
    if not semantics_check(T + r, ex):
      r = llm(sem_prmpt, T + r, ex)
      
    if regression_tests(T + r, reg_examples):
      T = T + r
      regression_examples.append(ex)
      break
\end{lstlisting}
\end{algorithm}

\noindent
We present our rule distillation method in Alg.~\ref{alg:ruledistill}.
Assume we want to extend an initial ASP theory $T$ and we have a distillation suite of new VQA examples at our disposal. Each example is a triple $(s,q,a)$ consisting of a scene description $s$ as ASP facts, a question description $q$ as ASP facts, and the expected answer $a$ as an ASP fact.

The algorithm iterates over all examples (line 5); for each example, it calls an ASP solver to check whether the expected answer can be derived from the theory accompanied with the scene and  question facts. If so, it skips to the next example (line 8); otherwise it tries to update the theory.

We apply different prompting strategies---more details follow below---to obtain a list of candidates for extending the theory. 
This involves one strategy in which we ask for several variants of a solution in one prompt, as well as two different chain-of-thought prompts. Typically, the quality of the resulting rules varies and they are not always correct. By synergetic use of the prompting strategies, we thus create a pool of options from which we will subsequently aim to pick one that works (lines 10--12). 

For each candidate rule (line 14), we use an ASP solver to deduce the answer for the current example: 
\begin{itemize}
\item[-] If we get a syntax error (line 15), we pass the error message to the LLM and prompt it to revise the rules (line 16); we call this step \emph{syntactic mending}. 

\item[-] If we get an answer set from the ASP solver, we check whether it matches the expected answer (line 18). If  not, 
we pass the answer 
to the LLM and ask it to update the rules (line 19); we call this step  \emph{semantic mending}.

\item[-] 
We then we do \emph{regression testing}, \iec testing whether a new rule appended to the theory gives us correct answers for all previously used examples (line 21).
This assures consistency and is in line with an online setting where future examples cannot be assumed to be known in advance.
Only after passing this step, we accept the suggested theory extension (line 22), add the example to the regression list (line 23) and break to the next example (line 24).
\end{itemize}

The models used tend to generally follow the instruction to return ASP rules, yet we observed that mistakes and hallucinations broadly fall into three categories: (i) syntactic, such as unsafe rules; (ii) semantic, where the generated rules fail to produce the expected answer for the current example; and (iii) regressive, where a rule may answer a specific question correctly but introduces a semantic conflict with previously distilled rules. These errors are systematically handled by our algorithm, using the ASP solver to ensure only syntactically valid and globally consistent rules are integrated into the theory.
We illustrate the three error types using the target primitive \texttt{verify\_attr(TO, TI, A, V)}, which checks whether an object at state $\texttt{TI}$ possesses an attribute $\texttt{A}$ with value $\texttt{V}$, and if so, state $\texttt{TO}$ is assigned ``yes''.

\begin{itemize}
    \item \textbf{(i) Syntactic Errors:} The model generates invalid ASP syntax, often hallucinating mathematical notation (\egc $\forall$, $\in$) symbols that are not supported by the solver.
    \begin{Verbatim}[fontsize=\small, commandchars=\\\{\}]
    bool(TO, yes) :- verify_attr(TO, TI, ATTR, VALUE), 
                     \(\forall\) ID \(\subseteq\) state(TI, ID), 
                     has_attr(ID, ATTR, VALUE).
    \end{Verbatim}

    \item \textbf{(ii) Semantic Errors:} The rule is syntactically valid but fails to produce the correct answer for the current example. In the case below, the model incorrectly requires the object to be ``burnt'' to satisfy the verification.
    \begin{Verbatim}[fontsize=\small]
    bool(TO, yes) :- verify_attr(TO, TI, ATTR, VALUE), 
                     state(TI, ID), 
                     has_attr(ID, ATTR, burnt).
    \end{Verbatim}

    \item \textbf{(iii) Regressive Errors:} The model generates a rule that works for the current example but conflicts with a previously distilled rule.
    
    \begin{Verbatim}[fontsize=\small]
    % Consider that the first input is an uncooked red lobster
    % The model distills the following rule
    bool(TO, no)  :- verify_attr(TO, TI, _, cooked), 
                     state(TI, ID), 
                     has_attr(ID, color, lobster).
    
    % Now consider the following input is a cooked red steak
    % The model distills the following rule
    bool(TO, yes) :- verify_attr(TO, TI, _, cooked), 
                     state(TI, ID), 
                     has_attr(ID, class, red).
    
    \end{Verbatim}
    Now the program derives yes (and no) when we present the uncooked red lobster, provoking a regression failure.
\end{itemize}

\subsection{Prompting}

We prompt the LLM in Alg.~\ref{alg:ruledistill} for different purposes: either to generate new rules or to revise rules if a syntax or semantics test failed.
For generating rules, we use different prompting strategies, but all extend the same preprompt that sets the stage. 
The prompts are rather general and only need small adjustments if  different VQA tasks are considered. 
Although it is well known that details of how prompts are formulated can have a great impact on the results, we are striving for clarity and readability instead of performance optimisation.
The complete prompts are presented in~\ref{apdx:prompts}.

\subsubsection{Preprompt}

Before prompting the LLM with VQA examples, we pass it a preprompt to prepare it for our rule distillation process. 
The preprompt instructs the LLM to only return ASP rules that extend an initial ASP theory. Most parts of the preprompt are dataset agnostic, but some, such as the scene, question, answer, and theory, are tailored to be suitable for each dataset. We illustrate preprompts and note whenever they are tailored for a specific dataset. 

A preprompt consists of the following parts, where each of them is illustrated with an excerpt of the complete preprompt:

\begin{enumerate}
    \item \textbf{Contextualisation.} We describe the VQA setting and clarify that we have already parsed both scene and question into ASP facts.
    \begin{Verbatim}[breaklines=true, breakanywhere=true]
We are in the domain of Visual Question Answering, where the problem
is to take an image and question related to it as input, and produce
as output the correct answer.
We have already preprocessed both the image and question into 
correct Answer Set Programming representations.
Scene/Question pairs of ASP facts serve as the instance for an ASP
program which we call the Theory.
    \end{Verbatim}
    \item \textbf{Logical representation.} We describe the syntax of the language that we use to represent questions and scenes, in our case ASP.
    \begin{Verbatim}[breaklines=true, breakanywhere=true]
Answer Set Programming (ASP) is a form of declarative programming 
oriented towards difficult search problems.
Its syntax and usage can be summarized as follows:

Rules: The basic building block of an ASP program. A rule has a head 
and a body, and is written in the form: Head :- Body. 
    \end{Verbatim}
    \item \textbf{Scene and question representation.} We explain the representation that is used for describing the scene and the question, respectively, and give examples (dataset tailored).
    \begin{Verbatim}[breaklines=true, breakanywhere=true]
Consider the following ASP representation for the objects 
in an image.
Object Declaration: Each object(<id>) declares a unique object 
with a specific identifier.
...
Consider the following ASP representation for natural language 
questions.
scene(S): Initializes all objects at step S.
\end{Verbatim}
    \item \textbf{Answer representation.} We describe the format of the answers to the questions (dataset tailored).
    \begin{Verbatim}[breaklines=true, breakanywhere=true]
The answer computed by the theory will always be in a predicate of
the form 'ans(X)', where 'X' is the expected answer.
'X' can be a yes or no, it can be an attribute, it can be a
relation or the value of some attribute.

Examples of these are:
ans(yes)
    \end{Verbatim}
    \item \textbf{Logical theory.} We present the initial theory $T$ that should be extended (dataset tailored)
    \begin{Verbatim}[breaklines=true, breakanywhere=true]
state(TO,ID) :- scene(TO), object(ID).
ans(V) :- end(TO), attr_value(TO,V).
ans(V) :- end(TO), attr(TO,V).
    \end{Verbatim}
    \item \textbf{Task explanation.} Finally, we explain the input that the LLM will receive from now on and the expected response.
    \begin{Verbatim}[breaklines=true, breakanywhere=true]
Your task is to keep the ASP theory updated with rules that allows
us to handle new questions.
Your will be presented with a VQA example containing a scene, 
question and expected answer.
Return ASP rules that are able to correctly process the example.    \end{Verbatim}
\end{enumerate}



While the distillation algorithm itself is domain-independent, certain components of the preprompt require manual adjustment when transitioning to a new dataset. Specifically, the engineering effort per dataset involves: (i) defining the scene representation, which describes how the symbolic solver views objects and their attributes; and (ii) defining the question representation, which outlines the available primitive operations and their expected arguments.
Each of these sections is accompanied by examples of how the ASP representations for scene and questions look like (see~\ref{apdx:prompts}).
We note that the majority of the preprompt structure remains static across tasks.

The preprompt will generally be followed by more detailed instructions of how to produce the new rules (\egc by using multi-prompting or chain-of-thought prompting strategies).

\subsubsection{Multi-prompting}

We use the strategy that instructs the LLM to produce not only one solution for the task, but to suggest several alternative solutions; we call this \emph{multi-prompting}. 
We apply this strategy in line 10 of Alg.~\ref{alg:ruledistill}. The advantage is that we can possibly obtain several solution candidates that we can evaluate using a single call to the LLM.
This increases the chances of finding a correct solutions without the need for several potentially expensive LLM calls. We use the following prompt:

{ 
\begin{Verbatim}[breaklines=true, breakanywhere=true]
  Consider you must output {N} different possibilities of the rules
  that need to be implemented.
  Separate the different answers by rows of three # symbols.
\end{Verbatim}
}

\noindent
The parameter \verb!N! is set to 5 by default.

\subsubsection{Chain-of-thought}

Chain-of-Thought (CoT) prompting is a technique in which an LLM is instructed to break down complex tasks into a series of intermediate steps~\citep{Wei0SBIXCLZ22}. This often helps the model to produce a more accurate and coherent response.
We adopted this idea for generating declarative rules, where we instruct the model to break down a problem solution as a set of if-then conditions and to translate them  in a subsequent step into ASP rules.
We use two versions of this idea in Alg.~\ref{alg:ruledistill}: a simple, rather straight-forward version (line 11), and a more intricate one that asks for more detailed explanations (line 12).
The simple version is as follows:
    
{ 
\begin{Verbatim}[breaklines=true, breakanywhere=true]
  You are required to produce a chain of thought. 
  Use the predicates in the theory to show how you arrive at the 
  correct solution.
  Only output one line per predicate used. 
\end{Verbatim}
}

The complex version of chain-of-thought prompting,
which is more verbose and structured than the previous, is as follows:

{ 
\begin{Verbatim}[breaklines=true, breakanywhere=true]
  You are required to produce a chain of thought. Follow the 
  steps below to break down domain knowledge and construct the rules:
  1. Understand the Problem
  2. Break Down the Domain Knowledge
  3. Formulate If-Then Statements
  4. Construct Logic Rules
\end{Verbatim}
}

\subsubsection{Prompts for rule mending}

If the suggested rules do not produce the expected results, we prompt the LLM to correct them.
We leverage feedback from the ASP solver to accomplish this. 
In particular, if a syntax test fails, we provide the LLM with the error message from the solver and a request to suggest a corrected version. Here is the prompt to achieve this:

{ 
\begin{Verbatim}[breaklines=true, breakanywhere=true]
You must repair the syntax of the prompted Answer Set Programming
rule(s). 
The solver has given the following error: {syntax_error}.
\end{Verbatim}
}

If the syntax is correct but we 
do not get the expected answer, we pass the answer from the ASP solver to the LLM and explain that we expected the correct answer instead. The prompt looks as follows:

{ 
\begin{Verbatim}[breaklines=true, breakanywhere=true]
The prompted rules are added to the theory do not produce the 
correct answer.
They do not calculate the correct answer, which is: {correct_answer},
but instead result in the following incorrect answer: {solver_answer}.
\end{Verbatim}
}

The semantic mending also includes the expected answer and the inferred answer, if any.

\section{Knowledge Distillation Experiments}\label{sec:experiments}

\begin{table}[ht]
\centering
\caption{Restoring the initial theory $\mathrm{T}$ of GQA after removing all rules mentioning predicate $P$.}
\label{tab:predicate_removal_GQA}
\begin{tabular}{lccccc}
\toprule
\textbf{Predicate} & \textbf{GPT-4o} & \textbf{DeepSeek} & \textbf{Mistral} & \textbf{LLaMA3} & \textbf{Gemini-3} \\
\midrule
query          & $97.44 \pm 2.48$ & $98.01 \pm 2.27$ & $98.86 \pm 0.08$ & $86.62 \pm 21.9$ & $98.92 \pm 0.00$ \\
exist          & $100.0 \pm 0.00$ & $99.98 \pm 0.01$ & $99.38 \pm 1.06$ & $89.58 \pm 6.01$ & $100.0 \pm 0.00$ \\
or             & $100.0 \pm 0.00$ & $100.0 \pm 0.00$ & $100.0 \pm 0.00$ & $94.27 \pm 1.67$ & $100.0 \pm 0.00$ \\
filter         & $97.24 \pm 3.38$ & $98.62 \pm 2.76$ & $100.0 \pm 0.00$ & $100.0 \pm 0.00$ & $100.0 \pm 0.00$ \\
choose\_attr   & $99.83 \pm 0.00$ & $99.79 \pm 0.06$ & $98.44 \pm 2.76$ & $99.74 \pm 0.07$ & $99.83 \pm 0.00$ \\
verify\_rel    & $98.94 \pm 1.23$ & $100.0 \pm 0.00$ & $94.70 \pm 2.52$ & $96.44 \pm 3.00$ & $100.0 \pm 0.00$ \\
select         & $100.0 \pm 0.00$ & $99.77 \pm 0.46$ & $99.53 \pm 0.38$ & $32.20 \pm 39.1$ & $100.0 \pm 0.00$ \\
negate         & $99.42 \pm 0.70$ & $100.0 \pm 0.00$ & $98.98 \pm 0.57$ & $99.04 \pm 0.59$ & $100.0 \pm 0.00$ \\
relate         & $98.13 \pm 2.69$ & $98.76 \pm 1.52$ & $99.11 \pm 1.54$ & $76.74 \pm 12.3$ & $100.0 \pm 0.00$ \\
two\_different & $100.0 \pm 0.00$ & $100.0 \pm 0.00$ & $100.0 \pm 0.00$ & $99.55 \pm 0.45$ & $100.0 \pm 0.00$ \\
two\_same      & $100.0 \pm 0.00$ & $100.0 \pm 0.00$ & $100.0 \pm 0.00$ & $98.76 \pm 0.00$ & $100.0 \pm 0.00$ \\
\bottomrule
\end{tabular}
\end{table}

\begin{table}[ht]
\centering
\caption{Restoring the initial theory $\mathrm{T}$ of CLEVR after removing all rules mentioning predicate $P$.}
\label{tab:predicate_removal_CLEVR}
\begin{tabular}{lccccc}
\toprule
\textbf{Predicate} & \textbf{GPT-4o} & \textbf{DeepSeek} & \textbf{Mistral} & \textbf{LLaMA3} & \textbf{Gemini-3} \\
\midrule
exist          & $100.0 \pm 0.00$ & $95.98 \pm 8.04$ & $100.0 \pm 0.00$ & $81.18 \pm 3.46$ & $100.0 \pm 0.00$ \\
unique         & $94.43 \pm 10.5$ & $100.0 \pm 0.00$ & --- & --- & $100.0 \pm 0.00$ \\
count          & $100.0 \pm 0.00$ & $100.0 \pm 0.00$ & $98.80 \pm 0.62$ & --- & $100.0 \pm 0.00$ \\
equal\_integer & $100.0 \pm 0.00$ & $98.21 \pm 3.10$ & $100.0 \pm 0.00$ & --- & $100.0 \pm 0.00$ \\
and            & $100.0 \pm 0.00$ & $94.65 \pm 2.74$ & $100.0 \pm 0.00$ & $95.79 \pm 2.98$ & $100.0 \pm 0.00$ \\
relate\_left   & $100.0 \pm 0.00$ & $93.14 \pm 8.70$ & $100.0 \pm 0.00$ & $89.82 \pm 7.20$ & $100.0 \pm 0.00$ \\
filter\_large  & $100.0 \pm 0.00$ & $100.0 \pm 0.00$ & $100.0 \pm 0.00$ & --- & $100.0 \pm 0.00$ \\
query\_shape   & $100.0 \pm 0.00$ & $100.0 \pm 0.00$ & $100.0 \pm 0.00$ & $88.01 \pm 6.92$ & $100.0 \pm 0.00$ \\
same\_color    & $100.0 \pm 0.00$ & $99.68 \pm 0.64$ & $100.0 \pm 0.00$ & $96.53 \pm 2.46$ & $100.0 \pm 0.00$ \\
\bottomrule
\end{tabular}
\end{table}

\begin{table}[ht]
\centering
\caption{Restoring the initial theory $\mathrm{T}$ of \graph after removing all rules mentioning predicate $P$.}
\label{tab:predicate_removal_CLEGR}
\begin{tabular}{lccccc}
\toprule
\textbf{Predicate} & \textbf{GPT-4o} & \textbf{DeepSeek} & \textbf{Mistral} & \textbf{LLaMA3} & \textbf{Gemini-3} \\
\midrule
cycle          & $99.19 \pm 0.00$ & $88.75 \pm 0.32$ & $87.44 \pm 2.31$ & --- & $99.78 \pm 0.00$ \\
shortest\_path & $91.73 \pm 6.07$ & $85.69 \pm 1.93$ & $85.10 \pm 1.16$ & --- & $100.0 \pm 0.00$ \\
paths          & $94.56 \pm 2.98$ & $80.74 \pm 7.45$ & --- & --- & $95.56 \pm 2.12$ \\
count\_nodes   & $91.91 \pm 8.52$ & $86.50 \pm 0.38$ & --- & --- & $100.0 \pm 0.00$ \\
adjacent       & $100.0 \pm 0.00$ & $78.90 \pm 7.30$ & $88.99 \pm 0.33$ & $81.29 \pm 1.96$ & $86.11 \pm 23.3$ \\
adjacent\_to   & $100.0 \pm 0.00$ & $74.74 \pm 11.8$ & --- & $88.59 \pm 0.57$ & $100.0 \pm 0.00$ \\
same\_line     & $98.76 \pm 2.34$ & $64.35 \pm 22.7$ & $87.14 \pm 0.20$ & --- & $100.0 \pm 0.00$ \\
common         & $98.60 \pm 0.53$ & $96.70 \pm 3.70$ & --- & --- & $98.22 \pm 0.50$ \\
line\_names    & $100.0 \pm 0.00$ & $87.62 \pm 4.22$ & --- & --- & $100.0 \pm 0.00$ \\
line\_count    & $100.0 \pm 0.00$ & $92.11 \pm 8.68$ & --- & --- & $100.0 \pm 0.00$ \\
exist          & $95.38 \pm 6.17$ & $93.80 \pm 3.10$ & --- & --- & $100.0 \pm 0.00$ \\
\bottomrule
\end{tabular}
\end{table}

\noindent
We conduct a series of experiments to evaluate our knowledge distillation method on the GQA, CLEVR, and \graph datasets to answer the following research questions:%
\footnote{The code for reproducing our experiments is available in an online repository: \url{https://github.com/pudumagico/KDASP}.} 

\begin{itemize}
\itemsep=3pt
    \item[]
{\em\bf (R1)}
Given an VQA task and correctly parsed examples of scenes, questions, and answers into ASP, can our approach extend the ASP theory to deal with questions that require operations/steps not yet encoded?

\item[]
{\em\bf (R2)}
Which LLMs are suitable for our method? 

\item[]
{\em\bf (R3)}
How do the individual prompting strategies contribute to the overall performance? 

\item[]
{\em\bf (R4)}
How do LLMs fare when prompted to compose new predicates from already existing ones?
\end{itemize}

Before going into the details of our experiments, we discuss the evaluation platform, the selection of distillation and testing instances from the datasets, and our selection of LLMs. 

\subsection{Evaluation Platform}
The evaluation platform is a workstation with an Intel Core i7-12700K CPU, 32GB of RAM, and an NVIDIA GeForce RTX 3080 Ti GPU with 12GB of video memory.
The ASP solver we use is \texttt{clingo}~\cite{gebser2019multi}, version 5.8.0, from the Potassco solver suite%
\footnote{\url{https://potassco.org/clingo/}}, an state-of-the-art system.
All experiments were run 5 times with a timeout of 600 seconds.
For reproducibility, we logged all our parameters, random seeds, input prompts, and resulting rules.

\subsection{Instances}\label{ssection:instances}
We create two sets of examples for each dataset, a \emph{distillation suite} for the distilling process of Algorithm~\ref{alg:ruledistill}, and a \emph{testing suite} to benchmark the distillation results. 
The suites were created aiming at having proportionally balanced examples of all primitive functions of each dataset.

For GQA, we used about $45k$ instances from its ``balanced training set'' and then divided this set into a distillation and test suites of about $35k$ and $10k$ instances, respectively.
For CLEVR, the splits are taken from its training set and contain $35k$ examples for distillation and $15k$ for testing.
For \graph, they also belong to the training set and contain $2k$ examples for distillation and $1k$ for testing.

\paragraph{Distillation Sample.}
Our preliminary experiments showed that the rules generated during distillation tend to be general, often covering multiple examples at once when a rule is distilled. 
To control resource usage and demonstrate the efficiency of our approach with limited data, we do not use the complete distillation suite at every step, but instead sample a small, fixed number of examples ($N = 10$) for each distillation attempt. While using larger sample sizes may improve performance by exposing more patterns, it is more costly and can also lead to deterioration due to increased prompt length and the resulting strain on the LLM’s reasoning capabilities. 
We have chosen this value empirically after our preliminary experiments, as it was large enough to produce rules without prohibitive cost.
On the other hand, for evaluation always the full test suite is used. 

\subsection{LLM Selection} 

The landscape of LLMs is rapidly evolving, with new models and variants emerging frequently, making it infeasible to cover all available options in our evaluation. We therefore focus on a representative subset of competitive and accessible models.

We ran experiments with several LLMs, both locally hosted and online, API-based ones. 
Small models that can be easily self-hosted showed unfortunately very poor performance for the considered rule distillation task. We tried in particular GPT4ALL\footnote{nomic.ai/gpt4all} ``wizardlm-13b'' and Zephyr 7b $\beta$\footnote{huggingface.co/HuggingFaceH4/zephyr-7b-beta}, which
is a fine-tuned version of the Mistral 7B model developed by the Hugging Face H4 team.  

Ultimately, we resort to larger API-based models. Our selection includes GPT-4o\footnote{\url{https://platform.openai.com/docs/models/gpt-4o}} from OpenAI, which is the company’s most advanced model, notable for its multi-modal capabilities (text, vision, and audio) and high alignment performance, accessible via a paid API service. In addition, we include three open-source models: DeepSeek\footnote{\url{https://huggingface.co/deepseek-ai/DeepSeek-V2-Chat}}, a flagship Chinese-English model designed for long-context reasoning and GPT-level performance; Mistral Large\footnote{\url{https://mistral.ai/news/mistral-large/}}, optimised for fast inference and general-purpose usage under an open-weight license (code not public but model weights are); Meta’s LLaMA 3 70B-Instruct\footnote{\url{https://huggingface.co/meta-llama/Meta-Llama-3-70B-Instruct}}, which offers strong instruction-following ability and high-quality outputs across tasks; and last, Gemini-3 Pro\footnote{\url{https://deepmind.google/models/gemini/pro/}}, which exhibits a new ``high reasoning'' mode that in practice translates to an embedded preprompt that makes the LLM ``think'' before answering.
In terms of scale, the models range from LLaMA 3 (70 billion parameters), Mistral Large ($\approx123$ billion)
DeepSeek-V2 (236 billion), and Gemini-3 Pro (estimated 1 trillion), up to GPT-4o, 
with estimated 1.8 trillion parameters.

The temperature parameter of an LLM regulates the randomness of its output: lower values (\egc near 0) make the model more deterministic by favouring high-probability tokens, while higher values introduce greater variability.
We set a temperature of 0 for the LLMs which theoretically assures deterministic behaviour.
Yet in practice, there are multiple sources of non-determinism when working with LLMs, such as different tokenisers~\cite{DBLP:conf/naacl/AliFTRLLKEDBJWJAJSOWSKF24}, floating point errors, and versions of systems under the same name which can evolve in the background.

\subsection{Theory Extension Simulation}\label{ssec:main_experiments}
For each VQA dataset, 
we start with the complete ASP theory $T$,
and remove all rules that mention a selected
predicate $P$ that occurs in some question representation.
Intuitively, $P$ represents the ASP implementation of some primitive function, as described in Section~\ref{sec:background}, which can be encoded in the theory using one or more rules.
We take a representative selection from the complete set of predicates and use samples as described in Subsection~\ref{ssection:instances} from the distillation suite, where each example selected must mention $P$. 
We then iterate over the examples using Algorithm~\ref{alg:ruledistill} to distil rules that can effectively calculate the expected answer.

The results for the datasets GQA, CLEVR, and \graph are shown in Tables~\ref{tab:predicate_removal_GQA},\ref{tab:predicate_removal_CLEVR}, and~\ref{tab:predicate_removal_CLEGR}, respectively.
The tables report, for each model and  predicate in the dataset, the average number of examples used ($n$) and the average accuracy (\%) achieved over the test suite, along with its standard deviation; the best results are in bold. 
The column labeled $\mathrm{T}\,{\setminus}\,P$ displays the accuracy of the initial ASP encoding with predicate $P$ removed, which serves as a baseline. 
The drop in accuracy depends on the number of questions affected and the role of 
predicate $P$; \egc \texttt{select} is used in almost every question of GQA as removing it drops the accuracy to 9.60\%. 
The other columns illustrate the performance of the evaluated LLMs; ``---'' indicates that there is no improvement, meaning, no new rules where produced.



A general overview per dataset allows one to see that for both GQA and CLEVR, LLMs were highly successful when distilling rules for them. 
On the other hand, \graph appears to be more challenging, as accuracies are in general inferior to the ones in other datasets, and some models could not produce an answer in some cases. 
This can be explained by the difficulty of the predicates involved in \graph, such as \texttt{cycle} and \texttt{shortest path}, in contrast to the focus on spatial and comparison relationships from GQA and CLEVR, which are more straightforward to implement. 
The predicates in \graph{} are usually larger in terms of number of rules used to encode them but are also more complex, as they involve recursion, while none of the rules for GQA nor CLEVR are recursive.

Analysing the results of every model reveals consistent patterns across datasets. 
GPT-4o remained the most reliable model overall, achieving near or perfect accuracy with low deviation, across most predicates and datasets. Notably on CLEVR, it was able to recover or produce semantically equivalent rules to the ones removed. 
DeepSeek showed strong performance as well, often trailing GPT-4o on GQA and CLEVR, but noticeably worse in \graph.

Mistral-Large demonstrated competitive results as well, being on par with DeepSeek on GQA and CLEVR (with the exception of the \texttt{unique} predicate in CLEVR where no rules were produced). Yet it followed the trend of deteriorating results for \graph, now some distillation attempts on particular predicates without improvement. 

Last, LLama3-70b's performance was acceptable for GQA, producing rules for almost every predicate, but with high variance and a low score on \texttt{select}, where the accuracy was only 32.20\%. 
For CLEVR, and in particular \graph, the results were noticeable worse; no new rules were produced for the distillation of many predicates. 

Gemini-3 demonstrated exceptional performance across all datasets, consistently achieving near-perfect accuracy (Tables~\ref{tab:predicate_removal_GQA}--\ref{tab:predicate_removal_CLEGR}). Notably, it achieved 100\% accuracy on several complex graph-based predicates in \graph, where other high-end models showed slight performance drops. 
We attribute this to the model's high-reasoning mode, which allows an LLM to (i) explore multiple reasoning paths, (ii) using ``verification loops'' to ensure that their reasoning is correct, and (iii) allocates a significant share of time to this preparatory reasoning step. In our setting, these characteristics are translated into a model that is more precise and robust than the rest.

The findings highlight that the distillation process used few examples and proved considerably successful for GPT-4o, DeepSeek, Mistral-Large and Gemini-3 on GQA and CLEVR, whereas Mistral-Large and LLaMA3-70b struggled on \graph where complex graph reasoning tasks posed greater difficulty.

Regarding the number of examples used, it always happened that fewer than all  available $N=10$ examples per distillation attempt were used. 
This is because the LLM may have already produced a set of rules general enough such that it can cover all future examples, or a set of rules was produced that the LLM could not improve upon. 
In general, the amount was remarkably low, using 3--8 examples to produce a set of rules.
While we set an upper limit of $N=10$ examples for our experiments, in practice, typically we use 1--4 examples per distillation attempt as shown in Table~\ref{tab:examples_used_summary}. 
The LLM often reaches a ``fixpoint'' during the process: once it produces a rule general enough to cover the representative logic of a question type, presenting further examples does not result in additional rules.
Gemini-3 was highly efficient, often requiring the fewest number of examples (averaging to only 1.73 for \graph) to converge on a correct set of rules.

\begin{table}[ht]
\centering

\caption{Average number of examples used by Model and dataset}
\label{tab:examples_used_summary}

\begin{tabular}{lccccc}
\toprule
\textbf{Dataset} & \textbf{GPT-4o} & \textbf{DeepSeek} & \textbf{Mistral} & \textbf{LLaMA3} & \textbf{Gemini-3} \\
\midrule
GQA   & $4.04 \pm 1.09$ & $2.16 \pm 0.32$ & $4.02 \pm 1.50$ & $4.03 \pm 0.67$ & $2.79 \pm 0.10$ \\
CLEVR & $3.22 \pm 0.50$ & $2.34 \pm 0.16$ & $3.90 \pm 1.29$ & $6.48 \pm 1.83$ & $2.58 \pm 0.32$ \\
\graph& $4.02 \pm 1.51$ & $2.40 \pm 0.32$ & $5.80 \pm 0.75$ & $5.02 \pm 0.24$ & $1.73 \pm 0.53$ \\
\bottomrule
\end{tabular}

\end{table}

For practitioners, the efficiency and operational cost of the distillation process are primarily determined by the token consumption of the LLMs. 
While input costs are largely a function of the preprompt size and the number of examples, output costs vary significantly based on the model's verbosity and its ability to adhere to the constraint of returning only ASP rules. Table~\ref{tab:tokens_used} provides a breakdown of the average input and output tokens per distillation attempt on 100 distillation attempts for each dataset using a random predicate. We observe that the results align with the number of examples needed per distillation attempt shown in Table~\ref{tab:examples_used_summary}, as models like DeepSeek and Gemini-3 use fewer tokens than the competitors. This is because their higher precision reduces the number of failed distillation attempts and repeated API calls triggered by the feedback loop. In contrast, other models like LLama often incur higher costs through multiple retries and exhibit high variance.

\begin{table}[ht]
\centering
\footnotesize
\caption{Average and standard deviation token usage per dataset and model}
\label{tab:tokens_used}

\begin{tabular}{lcccccc}
\toprule
 & \multicolumn{2}{c}{\textbf{GQA}} & \multicolumn{2}{c}{\textbf{CLEVR}} & \multicolumn{2}{c}{\textbf{\graph}} \\
\cmidrule(lr){2-3} \cmidrule(lr){4-5} \cmidrule(lr){6-7}
\textbf{Model} & \textbf{In} & \textbf{Out} & \textbf{In} & \textbf{Out} & \textbf{In} & \textbf{Out} \\
\midrule
GPT-4o        & $11259 \pm 762$ & $359\pm353$  
              & $6862 \pm743$ &  $356\pm42$
              & $3615 \pm336$ & $1931\pm 206$ \\
DeepSeek      & $8096 \pm 360$ & $319 \pm 24$ 
              & $3092 \pm 445$ & $793 \pm 130$ 
              & $3321 \pm 357$ & $325 \pm 458$ \\
Mistral-Large & $13591\pm 1494$ & $788\pm635$ 
              & $6551\pm532$ & $475\pm540$ 
              & $12604\pm1188$ & $1331\pm164$ \\
LLaMA3-70B    & $33565\pm2494$ &  $2587\pm 2048$
              & $25935\pm3007$  & $2942\pm3708$  
              & $39716\pm2574$ & $5166\pm4701$ \\
Gemini-3      & $8050 \pm 205$ & $128 \pm 40$ 
              & $3892 \pm 311$ & $82 \pm 36$ 
              & $4585 \pm 278$ & $368 \pm 214$ \\
\bottomrule
\end{tabular}
\end{table}

\subsection{Ablation Experiments}
Our distillation algorithm consists of a series of constituents including syntax mending, semantic mending, multi-prompting, two types of CoT and regression testing. 
We produce ablation studies focusing on three major elements of our prompting approach, which we refer to as \emph{distillation strategies}:
\begin{itemize}
\itemsep=3pt
\item \textbf{No Mending}: When removing both syntactic and semantic mending,
\item \textbf{No CoT}: When removing both simple and complex chain of thought prompts, and
\item \textbf{No Multiprompt}: When removing multi-prompting from the prompt.
\end{itemize}

Table~\ref{tab:avg_accuracy_std} reports the average accuracy of each model on each dataset with and without ablation. The best accuracy in each row is shown in bold; the best model without ablation for each dataset is additionally underlined. 
When a model failed to generate valid rules for a predicate, the accuracy was replaced with that of the theory $\mathrm{T} \setminus P$ as baseline.

Across datasets, GPT-4o and DeepSeek achieve consistently strong results. GPT-4o in particular benefits from the full prompting pipeline, especially on GQA and \graph: disabling any strategy in these settings leads to clear performance drops of  up to 10\%. 
DeepSeek shows similar behavior, though in CLEVR, disabling CoT slightly improves performance, indicating that simpler prompting can sometimes improve performance even on strong models.
In \graph, the impact of these strategies is generally less pronounced. 
Mistral-Large and LLama3-70b, which trail behind on GQA and \graph, perform comparably to larger models on CLEVR, especially when some strategies are deactivated. In these cases, reduced prompting appears to ease the burden on smaller models. 
For instance, LLama3-70b improves by 5\%--10\% on several CLEVR predicates when mending is disabled.
The same models exhibit high variance and inconsistent behaviour when distilling rules. 
LLama3-70b frequently fails to distill rules for harder tasks in \graph, regardless of strategy, and Mistral-Large sometimes fails completely for certain GQA or \graph predicates if strategies like multi-prompting are disabled. 
The ablation study for Gemini-3 reveals a remarkable stability compared to other models. While most models suffered significantly when mending or multi-prompting were disabled, Gemini-3 maintained an average accuracy of over 93\% even without mending. Unlike Mistral or LLaMA3, Gemini-3 was not hindered by complex prompting strategies, showing that its reasoning depth is sufficient to handle the full pipeline without being overburdened.

A more thorough analysis of the experiments is given in~\ref{apdx:ablation}.

The efficacy of distillation strategies is not uniform but rather highly correlated with the intrinsic reasoning capabilities of the underlying model. We observe three distinct performance archetypes: (i) \textit{Overburdened Models} (Mistral-Large, LLaMA3-70b), where the computational or context overhead of complex strategies like CoT and multi-prompting often outweighs their benefits, leading to better performance when these modules are deactivated; (ii) \textit{Balanced Models} (GPT-4o, DeepSeek), which effectively leverage these strategies to bridge reasoning gaps, showing clear performance gains when all distillation components are active; and (iii) \textit{Saturated Models} (Gemini-3), which possess such high baseline capabilities that they render some strategies redundant.


\begin{table*}[htb]
\centering
\caption{Average accuracy by model and ablation condition. For each dataset, best configuration per model is in bold, best model under full configuration is underscored.}
\label{tab:avg_accuracy_std}
\setlength{\tabcolsep}{2pt}
\begin{tabular}{llcccc}
\toprule
Dataset & Model         & Full & No Mending & No CoT & No Multi-prompt \\
\midrule

\multirow{4}{*}{GQA}
 & GPT‑4o        & $\mathbf{99.18 \pm 1.09}$ & $97.63 \pm 2.28$ & $98.06 \pm 1.59$ & $96.84 \pm 5.54$ \\
 & DeepSeek      & $\mathbf{99.54 \pm 0.72}$ & $98.97 \pm 1.54$ & $99.08 \pm 1.81$ & $99.36 \pm 1.32$ \\
 & Mistral‑Large & $\mathbf{99.00 \pm 1.53}$ & $97.94 \pm 2.91$ & $96.06 \pm 5.64$ & $98.74 \pm 2.19$ \\
 & LLama3‑70b    & $\mathbf{88.45 \pm 20.0}$ & $86.01 \pm 23.2$ & $82.98 \pm 8.43$ & $86.43 \pm 16.5$ \\
 & Gemini-3    & \underline{$\mathbf{99.89 \pm 0.32}$} & $98.89 \pm 0.59$ & $99.88\pm 0.52$ & $97.45 \pm 0.45$ \\

\midrule

\multirow{4}{*}{CLEVR}
 & GPT‑4o        & $\mathbf{99.38 \pm 1.86}$ & $98.43 \pm 4.72$ & $96.01 \pm 11.4$ & $96.85 \pm 9.44$ \\
 & DeepSeek      & $97.96 \pm 2.69$ & $98.61 \pm 2.59$ & $\mathbf{99.85 \pm 0.45}$ & $93.58 \pm 18.9$ \\
 & Mistral‑Large & $98.74 \pm 0.52$ & $95.07 \pm 14.0$ & $99.11 \pm 1.19$ & $\mathbf{99.21 \pm 0.51}$ \\
 & LLama3‑70b    & $94.07 \pm 6.33$ & $\mathbf{98.94 \pm 0.70}$ & $98.79 \pm 0.95$ & $87.91 \pm 22.5$ \\
 & Gemini-3    & \underline{$\mathbf{100.00 \pm 0.00}$} & $100.00 \pm 0.00$ & $100.00 \pm 0.00$ & $100.00 \pm 0.00$ \\
\midrule

\multirow{4}{*}{\graph}
 & GPT‑4o        & $\mathbf{97.28 \pm 3.28}$ & $86.86 \pm 4.14$ & $85.62 \pm 4.34$ & $88.34 \pm 1.61$ \\
 & DeepSeek      & $\mathbf{88.54 \pm 9.34}$ & $88.16 \pm 1.90$ & $86.98 \pm 3.27$ & $88.20 \pm 1.58$ \\
 & Mistral‑Large & $87.17 \pm 0.88$ & $\mathbf{87.25 \pm 1.44}$ & $84.25 \pm 8.37$ & $87.04 \pm 1.83$ \\
 & LLama3‑70b    & $\mathbf{86.76 \pm 1.87}$ & $86.33 \pm 3.14$ & $85.28 \pm 6.61$ & $85.39 \pm 4.10$ \\
 & Gemini-3    & \underline{$\mathbf{98.45 \pm 4.22}$} & $98.45 \pm 4.22$ & $97.51\pm1.31$ & $	98.12\pm4.75$ \\
\bottomrule
\end{tabular}
\end{table*}

\subsection{Composing New ASP Predicates}\label{ssection:composite}

We explore furthermore the synthesis of new ASP predicates by composing existing operations from a fixed ASP theory $T$, specifically on the GQA dataset. The primitives (\egc \texttt{select}, \texttt{has relation}, \texttt{filter}, \texttt{verify\_rel}) provide the foundational semantics for reasoning over visual scenes. 
Our goal is to test whether language models can define new predicates compositionally from those in $T$, in particular those with higher arity and those that use different ASP constructs.

We showcase how to introduce a new predicate, \texttt{in\_between}, which is a spatial predicate that relates three objects, and evaluate whether LLMs can provide its semantics using only the existing predicates in $T$.

The process unfolds in several steps:

\begin{enumerate}
\itemsep=3pt
    \item \textbf{Manual extension of the theory:} We begin by adding rules to handle a new predicate $P$ to $T$, such that it implements the desired predicate using predicates previously defined. For example, the predicate \texttt{in\_between} can be implemented as:
    \begin{verbatim}
state(TO, X) :- in_between(TO, T0, T1, T2),
                state(T1, ID1),
                state(T2, ID2),
                has_rel(X, to_the_right_of, ID1),
                has_rel(X, to_the_left_of, ID2).
    \end{verbatim}
    \item \textbf{Sampling modifiable examples:} We identify examples where the structure of the scene graph allows for a transformation into a query that involves the new predicate. These examples are modifiable into ones that speak about the desired predicate. For example, for the \texttt{in\_between} predicate, one searches for scenes with three or more objects. 
    \item \textbf{Transforming the examples:} Each selected example is transformed to
    \begin{itemize}
        \item a new ASP question using the new predicate (\egc \texttt{in\_between}),
        \item an updated answer, and
        \item an optionally modified scene if additional attributes or relations are needed.
    \end{itemize}

    \item \textbf{Dataset splitting:} We create distillation and testing suites.

    \item \textbf{Rule distillation:} To obtain new rules, we use Alg.~\ref{alg:ruledistill}, testing whether the LLM can produce $P$ by composing existing operations from $T$ using the generated examples as guidance.
\end{enumerate}

This method enables us to use a VQA dataset, in this case GQA, to probe the capabilities of LLMs to generate new predicates from exising ones.
Besides \texttt{in\_between}, we construct additional predicates to further evaluate predicate composition (their ASP encodings are given in~\ref{apdx:composite}):
\begin{itemize}
    \item For \texttt{connected}, we select two objects from the scene and ensure the presence of a path between them through one or more intermediate objects, \iec we synthesize relational chains by adding \texttt{has\_rel(A, connects, B)} relations (\egc object A is next to object B, and B is to the left of C) to form paths of length greater than two. The query then tests whether such indirect connections exist.
    \item For \texttt{isolated}, we identify an object and systematically remove all of its relationships from the scene graph, making it structurally disconnected. The associated query asks whether the object is isolated.
    \item For \texttt{count\_class}, we leave the scene unchanged and simply construct a query that counts the number of objects belonging to a given class. We note counting is not part of the original GQA primitives. 
\end{itemize}

\begin{table}[!t]
\caption{Distillation for producing new predicates removing all rules mentioning predicate $P$. Best average accuracy per predicate in bold.}
\label{tab:new_predicates_gqa}
\medskip
\centering
\footnotesize
 \setlength{\tabcolsep}{4pt}
\begin{tabular}{lccccc}
\toprule
$P$ & GPT-4o & DeepSeek & Mistral-Large & LLama3-70b & Gemini-3 \\ 
\midrule
in\_between & $97.39 \pm 2.82$ & $\mathbf{99.23 \pm 1.31}$ & $93.46 \pm 3.26$ & $92.82 \pm 3.20$ & $98.85 \pm 1.15$ \\
connected & $\mathbf{100.0 \pm 0.00}$ & $99.84 \pm 0.32$ & $99.92 \pm 0.16$ & $84.20 \pm 22.4$ & $\mathbf{100.0 \pm 0.00}$ \\
isolated & $\mathbf{100.0 \pm 0.00}$ & $\mathbf{100.0 \pm 0.00}$ & $\mathbf{100.0 \pm 0.00}$ & $74.09 \pm 26.3$ & $\mathbf{100.0 \pm 0.00}$ \\
count\_class & $\mathbf{100.0 \pm 0.00}$ & $\mathbf{100.0 \pm 0.00}$ & $\mathbf{100.0 \pm 0.00}$ & -- & $\mathbf{100.0 \pm 0.00}$ \\
\bottomrule
\end{tabular}
\end{table}

We use new distillation and test suites following the procedure described before, with $500$ examples each. 

For each new predicate, we report the average number $n$ of examples used, accuracy, and standard deviation, as shown in Table~\ref{tab:new_predicates_gqa}. 

GPT-4o can successfully provide the intended semantics for all novel predicates \texttt{connected}, \texttt{isolated}, and \texttt{count\_class} with perfect accuracy, the exception being \texttt{in\_between}, which achieves ca.\ $97\%$. 

Deepseek and Mistral both match GPT-4o on \texttt{isolated} and \texttt{count\_class}. 
The former even surpasses GPT-4o on \texttt{in\_between}, but at the cost of a lower than perfect score on \texttt{connected}. 
Mistral presents lower scores than DeepSeek on both \texttt{in\_between} and \texttt{connected}. LLama3-70b produced the worst results, in particular no rules were produced for \texttt{count\_class}.
Last, Gemini-3 positions itself again as the best performant model overall, achieving perfect accuracy on most of the predicates.

This experiment demonstrates the potential of LLMs to extend symbolic theories compositionally, a critical capability for scenarios where rules must be dynamically generated or adapted to new domains.

\subsection{Assessment of Research Questions} 
%

\smallskip
We now turn to our research questions R1--R4:

\smallskip
\noindent
{\bf (R1)} Given a VQA task and correctly parsed examples of scenes, questions, and answers
into ASP, can our approach extend the ASP theory to deal with questions that require
operations/steps not yet encoded?

Large language models can effectively recover missing ASP rules, with strong performance on GQA and CLEVR. Although performance drops on \graph, key primitive operations were still captured, promoting the viability of rule distillation across domains.

\smallskip
\noindent
{\bf (R2)} Which LLMs are suitable for our method?

Only large-scale models like GPT-4o, DeepSeek and Gemini-3 consistently succeed across tasks. Mistral-Large performs moderately well, while LLama3-70b struggles, showing that both scale and instruction-following capabilities are crucial for robust rule distillation.

\smallskip
\noindent
{\bf (R3)} How do the individual prompting strategies contribute to the overall performance?

Distillation strategies (mending, CoT, multi-prompting) substantially improve performance for top models, but can hinder weaker ones. GPT-4o and DeepSeek depend on them for complex cases, while Mistral and LLama3-70b often benefit from simpler prompting.

\smallskip
\noindent
{\bf (R4)} How do LLMs fare when prompted to compose new predicates from already existing ones?

GPT-4o, DeepSeek, and Gemini-3 can compose new ASP predicates using primitives from the base theory, even for recursive and aggregation-based rules. Mistral succeeds on simpler constructs, while LLama3-70b remains unreliable for compositional synthesis.

\smallskip

Regarding scalability with respect to the number of rules of the initial theory, we observe that the distillation algorithm operates locally by processing one primitive predicate at a time. Consequently, the total size of the theory has low impact on the effectiveness of it.
In fact, larger initial theories may improve performance by providing the LLM with richer context to prevent reasoning drift, whereas extremely sparse theories risk insufficient context.
In terms of reasoning capabilities, our experiments across GQA, CLEVR, and \graph demonstrate robustness in spatial, equality, counting, and recursive graph-related tasks, covering a vast landscape of ASP constructs. A future direction is to extend this approach to temporal datasets, leveraging ASP's native temporal extensions\footnote{\url{https://github.com/potassco/telingo}} to handle time-dependent predicates within the VQA encoding.

In summary, our results show that LLMs can effectively recover missing ASP rules in the context of VQA and extend symbolic theories with new logical rules. 
Success depends heavily on both dataset complexity and model capabilities: simpler tasks in GQA and CLEVR are addressed with high accuracy, while structurally demanding domains like \graph, which involve recursion and aggregation, require larger models and robust prompting strategies.



\section{Pruning Extracted Rules}\label{sec:exrul}
It happened that many logic programs produced in our experiments include \emph{redundant rules}---\iec rules whose removal does not affect the answer of the queried predicates. 
Redundant rules can arise from overly specific rule instantiations or responses by LLMs where a correct rule is produced alongside other innocuous ones which do not interfere with the correct computation.

Redundancy increases the complexity of rule sets, makes analysis and debugging harder, and may obscure the core reasoning pattern. Furthermore, in neurosymbolic systems where extracted rules are passed back into neural architectures (\egc for neurosymbolic learning or data augmentation), compactness and non-redundancy are crucial for interpretability and computational efficiency.

We now illustrate on output of our distillation algorithm how pruning a selection of the produced rules may simplify extracted rule sets without loss of accuracy with respect to some distillation data. 

Consider an instance of rules distilled for the \texttt{select} predicate in the GQA dataset using GPT-4o:

\begin{Verbatim}[breaklines=true, breakanywhere=true]
state(TO,ID)   :- select(TO, TI, CLASS), state(TI, ID), 
                  has_attr(ID, class, CLASS).
state(TO,ID)   :- select(TO, TI, CLASS), unique(TO, TI), state(TI, ID), 
                  has_attr(ID, class, CLASS).
is_attr_value(ID, X) :- query(TO, TI, ATTR), state(TI, ID), 
                        has_attr(ID, ATTR, X).
state(TO,ID)   :- select(TO, TI, CLASS), state(TI, ID), 
                  has_attr(ID, class, CLASS).
\end{Verbatim}

This example contains multiple rules produced by the LLM during the distillation process that are not useful to solve the \textit{select} query. After eliminating the redundant rules, the set reduces to:

\begin{Verbatim}[breaklines=true, breakanywhere=true]
state(TO,ID) :- select(TO, TI, CLASS), state(TI, ID), 
                has_attr(ID, class, CLASS).
\end{Verbatim}

The pruned theory achieves 100\% accuracy on the GQA distillation suite, equal to the accuracy without pruning

\subsection{Rule Redundancy Heuristic}\label{sec:redundancy}


We present our redundancy elimination procedure in Algorithm~\ref{alg:redundancy}. 
Assume we have the results of a distillation process for a predicate $P$: a set $R$ of new ASP rules produced by the LLM, the initial theory $T \setminus P$, and a pruning sample drawn from the distillation suite. 

Since rules generated later in Alg.~\ref{alg:ruledistill} tend to be more general due to regression testing, as new distilled rules must capture past examples as well, we attempt to eliminate rules from first to last.
The algorithm iterates over all rules $r_i \in R$ in the order they were generated (line~7).

For each rule, it temporarily removes $r_i$ to form a reduced set $Q$ (line~8), and calls \texttt{evaluate} to compare the behaviour of extending $T$ with both sets on the pruning sample (line~9). 
If the outputs are identical, the rule is marked as redundant (line~10). 
After all candidates have been checked, the redundant rules are removed from $R$ (line~12).

\begin{algorithm}[t]
\caption{Rule redundancy elimination.}
\label{alg:redundancy}
\begin{lstlisting}[language=Python]
# T\P: initial theory
# R: list of new ASP rules produced by the LLM
# pruning_sample: sample of training scenes, questions, and answers

redundant_rules = []

for ri in R:
 Q = R - {ri}
 if evaluate(T\P + R, pruning_sample) == evaluate(T\P + Q, pruning_sample):
   redundant_rules.append(ri)

R = R - set(redundant_rules)
\end{lstlisting}
\end{algorithm}

We use a sample from the examples for distillation also for pruning.
If a rule can be removed without affecting correctness on this sample, it is marked as redundant and discarded. 

\begin{figure*}[!htbp]
    \centering

    \begin{subfigure}{0.32\textwidth}
        \centering
        \includegraphics[width=\textwidth]{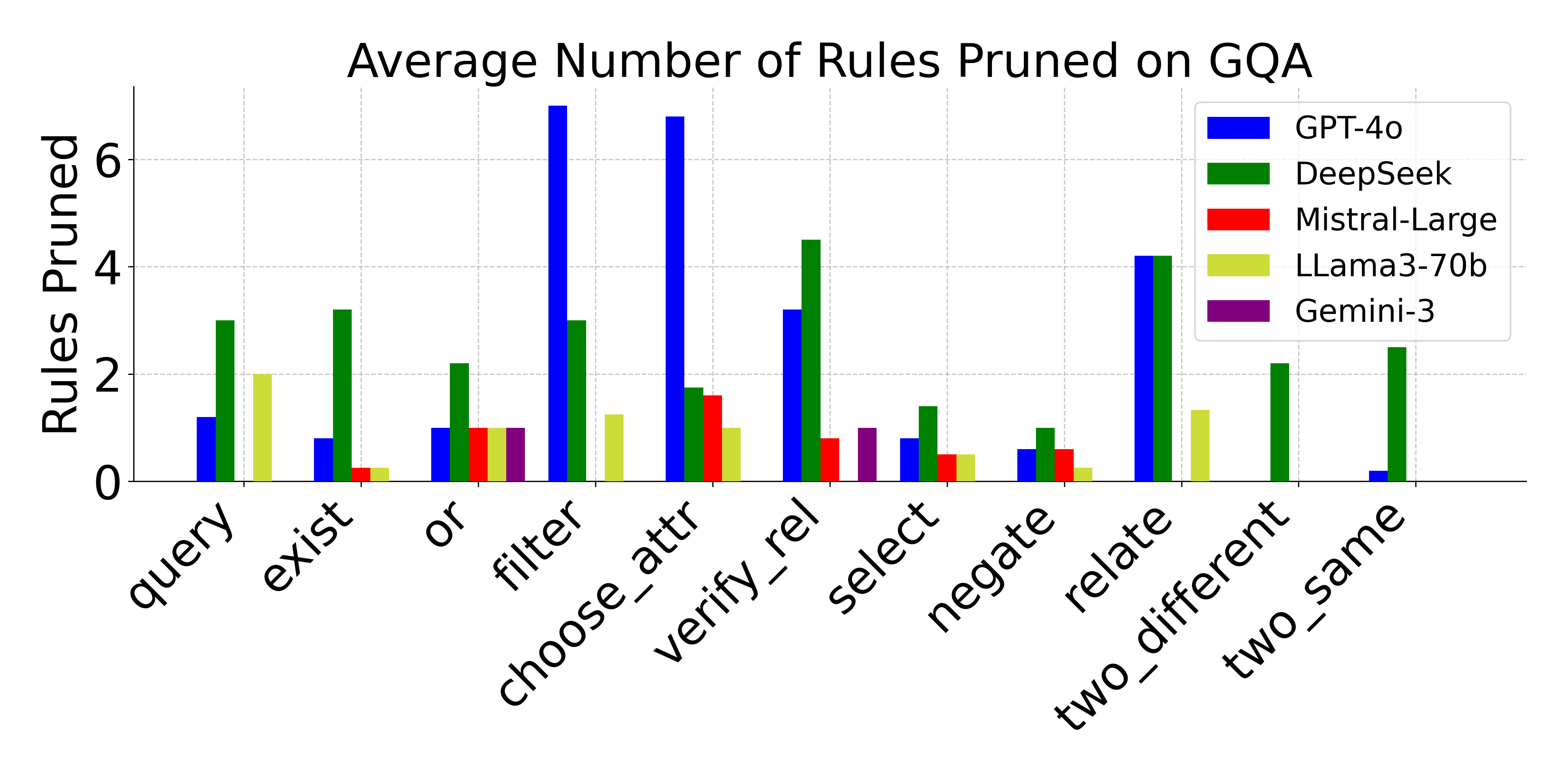}
        \caption{GQA.}
        \label{fig:pruning_GQA}
    \end{subfigure}
    \hfill
    \begin{subfigure}{0.32\textwidth}
        \centering
        \includegraphics[width=\textwidth]{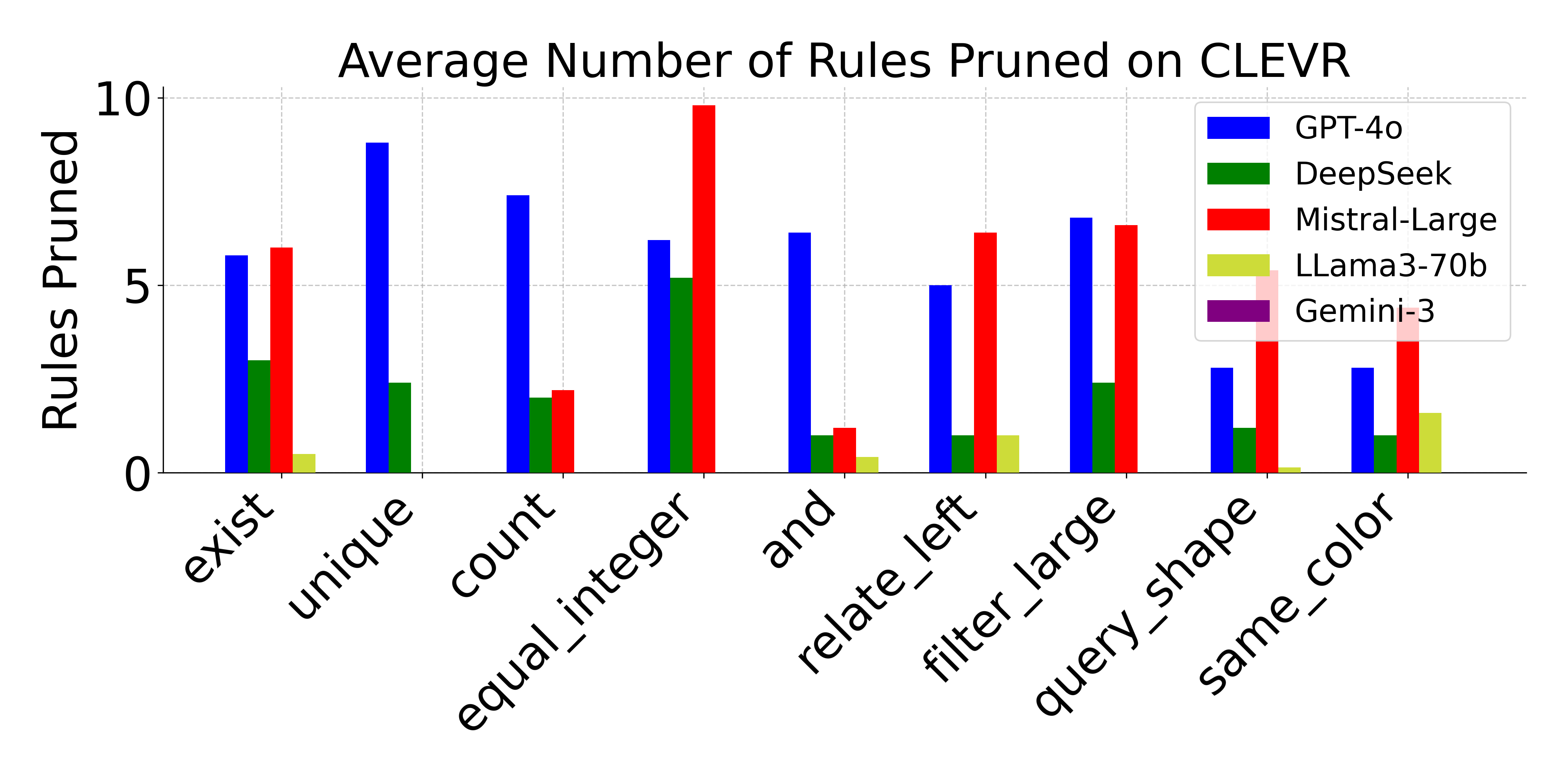}
        \caption{CLEVR.}
        \label{fig:pruning_CLEVR}
    \end{subfigure}
    \hfill
    \begin{subfigure}{0.32\textwidth}
        \centering
        \includegraphics[width=\textwidth]{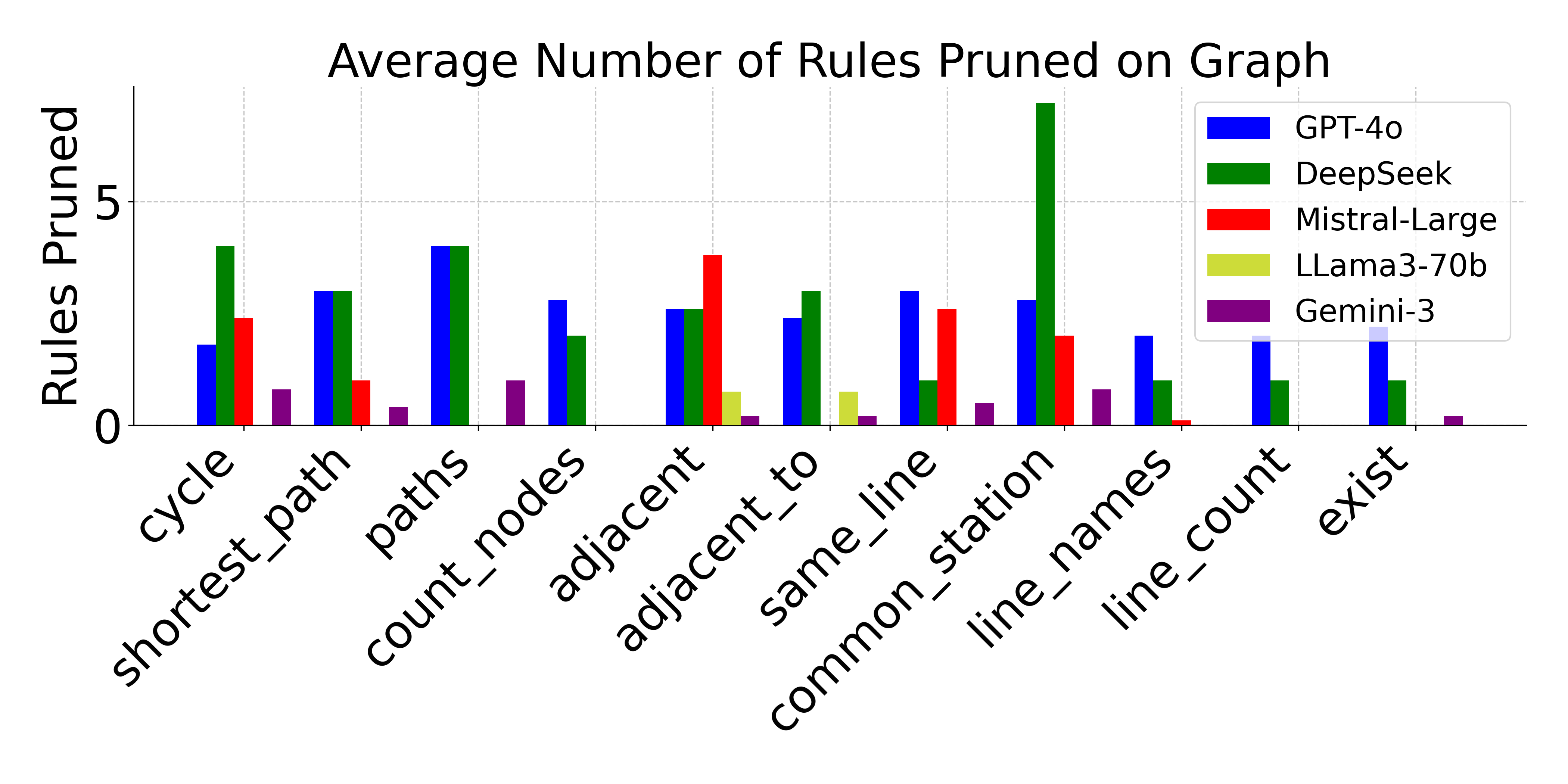}
        \caption{\graph.}
        \label{fig:pruning_CLEGRV}
    \end{subfigure}

    \caption{
    Average number of rules pruned for each predicate across different models on the GQA, CLEVR, and \graph datasets.
    }
    \label{fig:pruning_all_datasets}
\end{figure*}

Fig.~\ref{fig:pruning_all_datasets} visualises the outcomes of applying Algorithm~\ref{alg:redundancy} on the new rules generated in the main experiments (Section~\ref{ssec:main_experiments}) for GQA, CLEVR, and \graph; values in tabular form are presented in~\ref{apdx:pruning}. 
For each predicate, the bars indicate the average number of rules pruned across the sets of new rules produced by the LLM. 
We use a sample of 1000 examples from the corresponding distillation suite of each predicate.

Across all datasets, pruning reveals meaningful differences in how LLMs approach rule generation. GPT-4o consistently produces more extensive rule sets, often leading to high pruning counts---such as 8.8 rules for \textit{unique} in CLEVR---suggesting a thorough but sometimes redundant exploration of possible solutions.
In contrast, Mistral-Large and LLama3-70b tend to yield more compact outputs, with near-zero pruning for several predicates, such as \textit{relate} or \textit{select} in GQA. 
Gemini-3 exhibited the lowest pruning rates. In many cases, it produced zero redundant rules. This indicates that the model ``high'' reasoning mode allows to target the exact logic required by the VQA primitive as an ASP rule.

Low pruning, combined with high accuracy and a low number of examples, indicates that the LLM has produced general and useful rules. 
In contrast, high pruning with many examples suggests exploratory behaviour, where the LLM initially generates a broader set of rules that are gradually refined through the distillation process. 
However, when high pruning occurs despite few examples being used, it may indicate that the LLM is generating superfluous rules—statements that do not affect the outcome.

GQA and CLEVR present the highest pruning rates overall, as their simpler and more compositional predicates encourage LLMs to generate varied candidate rules. 
In contrast, \graph shows consistently lower pruning counts across models---not due to more efficient generation, but because its increased complexity makes rule synthesis harder in the first place, often resulting in fewer or no rules to prune. 
The pruned rules---alongside observed syntactic or semantic errors---offer possibly valuable data that can be potentially incorporated as prompt feedback or leveraged as training data to fine-tune models toward more accurate rule generation.

\section{Related Work}\label{sec:rel}

Our work intersects several areas of research that integrate symbolic reasoning with machine learning. In the following, we review relevant developments in ASP and LLMs, statistical-relational learning, and modular neurosymbolic approaches to VQA.

\subsection{Answer-Set Programming and LLMs}
Recently, LLMs have been explored in the context of ASP. \cite{ishay2023llmasp} observed that LLM reasoning capabilities are shallow, but they can serve as a highly effective semantic parser to transform input into ASP representations. These are then used to solve logical puzzles.
Furthermore, \cite{bauer24} recently proposed to use LLMs to parse questions into ASP facts in the context of VQA for images of graphs.

\cite{gupta2023reliable} showed that the combination of LLMs with ASP in their STAR framework produces good results for natural language tasks that require qualitative reasoning, mathematical reasoning, and goal-directed conversation. 
\cite{DBLP:conf/ijcai/SantanaKR24} proposed a two-step system for translating natural language to ASP via a controlled natural language intermediate form, demonstrating accurate rule generation on graph-based tasks.
LLASP~\citep{DBLP:conf/kr/CoppolilloC0PR24} is an approach that involves fine-tuning LLMs specifically for ASP, showing that model adaptation can significantly improve rule generation accuracy in logic programming tasks.

Our work is different as we do not focus on semantic parsing but on the more challenging task of knowledge distillation, where we aim for a system that can produce sound logical rules capturing knowledge about a particular domain.

Recent work by \cite{zhu2023llm} explores the use of LLMs to learn rules from arithmetic and kinship relationships, yet the rules they learn do not contain variables and their semantics is informal.

\subsection{Statistical-Relational Learning}
Similar to our approach, methods from statistical-relational learning (SRL)~\citep{RaedtK17}, in particular from inductive logic programming (ILP)~\citep{DBLP:journals/ngc/Muggleton91,MuggletonR94,DBLP:journals/ml/CropperDEM22}, aim to produce rules from example data and a background theory.
SRL has seen great advances in terms of scalability by, \egc applying gradient-based boosting~\citep{GutmannK06}, and systems like ILASP~\citep{law20} and FastLAS~\citep{LawRBB020} provide means for inductively learning expressive ASP programs.  \cite{DBLP:conf/lpnmr/KareemGBRR24} combine LLMs with ILASP to induce commonsense rules from answer sets for robust QA, requiring only few examples. 

SRL adopts a statistical and probabilistic perspective that is fundamentally data-driven. In contrast, our approach does not perform learning.
While we also make use of a dataset, its role is to guide a knowledge distillation process by conditioning an LLM, rather than to serve as training data. Notably, only a few examples are sufficient for this purpose.

Traditional ILP uses a search-based approach where the solutions produced are correct and minimal under some criteria. A key aspect of many ILP systems are {\em mode declarations} that define the syntactic form of allowed rules to restrict the search space of possible programs.
They are given in a
formal language and tacitly assume an intuition about the form of the solution.
The distillation approach does not require this; we only elicit knowledge that is already present in the LLM, and the information in the prompt that instructs what rules we want is informal and in natural language. When prompting LLMs, rules are general by command---while optimality is not enforced, it may happen implicitly.
In very recent work, \cite{DBLP:journals/corr/abs-2505-21486} propose to use LLM agents to generate the language bias itself, enabling automated hypothesis generation in ILP.

\subsection{Modular Neurosymbolic VQA}
There are several VQA systems that feature a modular architecture that combines subsymbolic with symbolic components~\citep{yi2019neuralsymbolicvqa,MaoGKTW19,amizadeh2020neurosymbolic,eiter2022neurosymbolicasp,vipergpt,JohnstonNS23a}.

Specifically, \cite{yi2019neuralsymbolicvqa} used a pipeline to extract a \emph{scene graph} (a list of all objects detected in the image with their attributes and positions) from the image. They then translated the provided question into a structured representation of the reasoning steps, called \emph{functional program}, and executed this program on the structural scene representation to obtain an answer. The authors showed excellent results on the popular CLEVR dataset~\citep{johnson2017clevr}. 
This approach has been advanced with logic-based reasoning processes, \egc by Differentiable First-Order Logic ($\nabla$-FOL)~\citep{amizadeh2020neurosymbolic}, or by ASP~\citep{eiter2022neurosymbolicasp}. 
These reasoning processes 
can consider not just the most probable scene-graph prediction, but rather the entire vector of probabilities as output by the object detection and attribute/relation classifier networks that form the visual perception component of the VQA pipeline.

While our focus is on VQA, related developments exist in textual QA. For instance, \cite{DBLP:conf/icaart/NguyenPVTN25} use LLMs in combination with ASP to detect misleading content by generating logic-based explanations. Their system highlights parts of a text responsible for incorrect or hallucinated answers, illustrating the potential of symbolic reasoning for enhancing transparency of QA pipelines.

Foundational models including Vision-Language Models (VLMs), such as BLIP-2~\citep{li2023blip2} and SimVLM~\citep{wang2022simvlm}  have become sufficiently strong through their pre-training regimes to generalise well to multiple different datasets. Approaches that use these VLMs as components are, \egc ViperGPT~\citep{vipergpt}, CodeVQA~\citep{subramanian2023codevqa}, and PnP-VQA~\citep{tiong2023pnpvqa}. 

Regarding systems for neurosymbolic learning, it should be mentioned that there are logic-based systems that employ the semantic loss~\citep{xu2018sl} to use the information produced by the reasoning module to improve the learning tasks of the neural networks involved~\citep{yang2020neurasp,ManhaeveDKDR21}.

\section{Conclusion}\label{sec:concl}

We have presented a method for declarative knowledge distillation using large language models (LLMs), which serves to generate rules for the reasoning component of a visual question answering (VQA) system. By leveraging examples from a dataset as guidance and relying exclusively on prompting—without requiring fine-tuning—the approach enables an automated extension of answer-set programming (ASP) theories. We have demonstrated the usefulness of the method across diverse datasets; while designed for VQA, it is in principle generic and could be explored in other contexts as well.

Combining logic-based reasoning with neural models provides systems that are interpretable and verifiable, properties that are crucial for trustworthy AI. 
Logical rules are an intuitive and transparent basis for encoding complex reasoning behaviour and can facilitate explainability in neurosymbolic architectures. 
Our results show that LLMs can effectively synthesise ASP rules to extend reasoning capabilities in VQA, a task that requires understanding and manipulating symbolic representations.
Among the evaluated models, GPT-4o and DeepSeek outperform others, particularly when enhanced with distillation strategies such as multi-prompting and chain-of-thought (CoT) prompting. 
Furthermore, Gemini-3 Pro demonstrated exceptional capability, consistently achieving near-perfect accuracy across all datasets while producing the most compact rule sets. This superior performance is attributed to the model's "high reasoning mode," which allows the model to internally explore and validate logical paths before generating the output ASP rule(s).
Notably, the ability of LLMs to propose rule completions and revisions indicates that they could serve as interactive \emph{copilots}, supporting practitioners in the development of logical programs.

Despite its promise, the proposed method has several limitations. It assumes the availability of datasets with decomposable, compositional question semantics, which may restrict its applicability in domains lacking such structure. 
Prompt sensitivity remains a challenge, though our templates demonstrated robust performance across different datasets. Furthermore, rule correctness is validated empirically rather than formally, although constraints and regression tests mitigate some potential errors.

Future work includes applying prompt tuning to better specialise LLMs for ASP rule generation, aiming to improve accuracy and reduce example requirements. 
We also plan to extend our distillation method beyond VQA to dynamic domains (\egc biomedical knowledge, real-time monitoring, learning agents) and structured question answering, where symbolic inputs and multi-step reasoning are central.

\section*{Acknowledgments}

This work was partially funded from the Bosch Center for AI.

{ 
\bibliographystyle{abbrvnat}
\bibliography{ref.bib}
}

\newpage
\appendix

\section*{Appendix}

The appendix provides additional resources, examples, and details that complement the main body of the article. It is organized as follows:

\begin{itemize}
    \item \textbf{Appendix A} contains examples of the ASP-based scene and question representation, along the full ASP theories used in the reasoning modules for each of the three VQA datasets: GQA, CLEVR, and \graph. These include the complete set of logic rules for primitive functions.
    
    \item \textbf{Appendix B} contains the full prompts used in our distillation procedure, including the preprompts and specific prompting strategies such as multi-prompting, chain-of-thought, and rule mending.
    
    \item \textbf{Appendix C} expands the ablation analysis with per-predicate performance breakdowns across all strategies and models.
    
    \item \textbf{Appendix D} lists the ASP encodings of the newly synthesized predicates used in the compositional generalisation experiments.
    
    \item \textbf{Appendix E} includes full pruning statistics for all predicates and models.
\end{itemize}

\section{ASP Theories}~\label{apdx:theories} 
This section includes complete ASP encodings used for our experiments. For each dataset—GQA, CLEVR, and \graph—we show (i) representative question programs in ASP format, (ii) the corresponding scene encodings, and (iii) the full ASP theories, including primitive operations and rules used for evaluation.
\subsection{GQA}
\medskip

\paragraph{GQA Question Encoding}

{ \small
\begin{Verbatim}[breaklines=true, breakanywhere=true]
scene(0).
select(1, 0, meat).
filter(2, 1, color, brown).
unique(3, 2).
verify_attr(4, 3, any, cooked).
scene(5).
select(6, 5, chair).
relate(7, 6, bacon, to_the_left_of, subject).
filter_any(8, 7, burnt).
unique(9, 8).
verify_attr(10, 9, any, cooked).
and(11, 4, 10).
end(11).
\end{Verbatim}
}

\paragraph{GQA Scene Encoding}

{ \small
\begin{Verbatim}[breaklines=true, breakanywhere=true]
object(1243751).
has_attr(1243751, class, bread).
has_attr(1243751, name, bread).
has_attr(1243751, class, baked_good).
has_attr(1243751, tone, light).
has_attr(1243751, weight, light).
has_attr(1243751, any, crispy).
has_attr(1243751, any, toasted).
has_attr(1243751, hposition, middle).
has_attr(1243751, vposition, middle).
has_rel(1243751, above, 1243746).
object(1243750).
has_attr(1243750, class, finger).
has_attr(1243750, name, finger).
has_attr(1243750, color, white).
has_attr(1243750, hposition, left).
has_attr(1243750, vposition, top).
has_rel(1243750, touching, 1243745).
object(1243746).
has_attr(1243746, class, egg).
has_attr(1243746, name, egg).
has_attr(1243746, class, food).
has_attr(1243746, any, cooked).
has_attr(1243746, any, bright).
has_attr(1243746, color, yellow).
has_attr(1243746, hposition, middle).
has_attr(1243746, vposition, middle).
has_rel(1243746, below, 1243751).
object(1243747).
has_attr(1243747, class, bacon).
has_attr(1243747, name, bacon).
has_attr(1243747, class, food).
has_attr(1243747, class, meat).
has_attr(1243747, color, dark).
has_attr(1243747, tone, dark).
has_attr(1243747, any, cooked).
has_attr(1243747, any, burnt).
has_attr(1243747, hposition, left).
has_attr(1243747, vposition, bottom).
has_attr(1243747, vposition, middle).
has_rel(1243747, to_the_left_of, 3827420).
object(1243745).
has_attr(1243745, class, sandwich).
has_attr(1243745, name, sandwich).
has_attr(1243745, class, food).
has_attr(1243745, color, brown).
has_attr(1243745, width, wide).
has_attr(1243745, size, large).
has_attr(1243745, any, seasoned).
has_attr(1243745, hposition, middle).
has_attr(1243745, vposition, middle).
has_rel(1243745, with, 1243748).
object(3827418).
has_attr(3827418, class, cheese).
has_attr(3827418, name, cheese).
has_attr(3827418, class, food).
has_attr(3827418, class, flavor).
has_attr(3827418, hposition, middle).
has_attr(3827418, vposition, middle).
object(3827419).
has_attr(3827419, class, person).
has_attr(3827419, name, person).
has_attr(3827419, class, person).
has_attr(3827419, hposition, right).
has_attr(3827419, vposition, bottom).
has_attr(3827419, vposition, middle).
has_rel(3827419, sitting_in, 3827420).
object(3827420).
has_attr(3827420, class, chair).
has_attr(3827420, name, chair).
has_attr(3827420, class, furniture).
has_attr(3827420, hposition, right).
has_attr(3827420, vposition, bottom).
has_attr(3827420, vposition, middle).
has_rel(3827420, to_the_right_of, 1243747).
object(1243748).
has_attr(1243748, class, meat).
has_attr(1243748, name, meat).
has_attr(1243748, class, food).
has_attr(1243748, class, meat).
has_attr(1243748, any, cooked).
has_attr(1243748, color, brown).
has_attr(1243748, hposition, middle).
has_attr(1243748, vposition, middle).
has_rel(1243748, inside, 1243745).
object(1243749).
has_attr(1243749, class, hand).
has_attr(1243749, name, hand).
has_attr(1243749, hposition, left).
has_attr(1243749, vposition, top).
\end{Verbatim}
}

\paragraph{GQA ASP theory}

{ \small
\begin{Verbatim}[breaklines=true, breakanywhere=true]
state(TO,ID) :- scene(TO), object(ID).
state(TO,ID) :- select(TO, TI, CLASS), state(TI, ID), 
                has_attr(ID, class, CLASS).
state(TO,ID) :- filter(TO, TI, ATTR, VALUE), state(TI, ID), 
                has_attr(ID, ATTR, VALUE).
state(TO,ID) :- filter_any(TO, TI, VALUE), state(TI, ID), 
                has_attr(ID, ATTR, VALUE).
state(TO, ID') :- relate(TO, TI, CLASS, REL, subject), state(TI, ID), 
                  has_attr(ID', class, CLASS), has_rel(ID', REL, ID).
state(TO, ID') :- relate(TO, TI, CLASS, REL, object), state(TI, ID),                             has_attr(ID', class, CLASS), has_rel(ID, REL, ID').
state(TO, ID') :- relate_any(TO, TI, REL, subject), state(TI, ID),
                  has_rel(ID', REL, ID).
state(TO, ID') :- relate_any(TO, TI, REL, object), state(TI, ID), 
                  has_rel(ID, REL, ID').
state(TO, ID') :- relate_attr(TO, TI, CLASS, ATTR), state(TI, ID), 
                  has_attr(ID, ATTR, VALUE), has_attr(ID', class, CLASS), 
                  has_attr(ID', ATTR, VALUE'), VALUE==VALUE', ID!=ID'.
{ has_attr(ID, ATTR, VALUE) 
: is_attr_value(ID, VALUE)} = 1 :- query(TO, TI, ATTR), state(TI, ID), 
                                   ATTR != name, ATTR != class, 
                                   ATTR != hposition, ATTR != vposition.
attr_value(TO,VALUE) :- query(TO, TI, ATTR), state(TI, ID), 
                        has_attr(ID, ATTR, VALUE).
bool(TO, yes) :- verify_attr(TO, TI, ATTR, VALUE), 
                 state(TI, ID), has_attr(ID, ATTR, VALUE).
bool(TO,no)   :- verify_attr(TO, TI, ATTR, VALUE), not bool(TO,yes).
bool(TO, yes) :- verify_rel(TO, TI, CLASS, REL, subject), state(TI, ID),
                 has_attr(ID', class, CLASS), has_rel(ID', REL, ID).
bool(TO,no)   :- verify_rel(TO, TI, CLASS, REL, subject), not bool(TO,yes).
bool(TO, yes) :- verify_rel(TO, TI, CLASS, REL, object), state(TI, ID), 
                 has_attr(ID', class, CLASS), has_rel(ID, REL, ID').
bool(TO,no)   :- verify_rel(TO, TI, CLASS, REL, object), not bool(TO,yes).
{has_attr(ID, ATTR, VALUE) ; 
 has_attr(ID, ATTR, VALUE')} = 1 :- choose_attr(TO, TI, ATTR, VALUE, VALUE'),         
                                    state(TI, ID).
attr_value(TO, VALUE)  :- choose_attr(TO, TI, ATTR, VALUE, VALUE'), 
                          state(TI, ID), has_attr(ID, ATTR, VALUE).
attr_value(TO, VALUE') :- choose_attr(TO, TI, ATTR, VALUE, VALUE'), 
                          state(TI, ID), has_attr(ID, ATTR, VALUE').
{has_rel(ID', REL, ID) : 
 has_attr(ID', class, CLASS) ; 
 has_rel(ID', REL', ID) :
 has_attr(ID', class, CLASS)} = 1 :- choose_rel(TO, TI, CLASS, 
                                                REL, REL', subject), 
                                     state(TI, ID).
rel(TO, REL)  :- choose_rel(TO, TI, CLASS, REL, REL', subject), state(TI, ID), 
                 has_attr(ID', class, CLASS), has_rel(ID', REL, ID).
rel(TO, REL') :- choose_rel(TO, TI, CLASS, REL, REL', subject), state(TI, ID), 
                 has_attr(ID', class, CLASS), has_rel(ID', REL', ID).
{has_rel(ID, REL, ID') :
 has_attr(ID', class, CLASS) ;
 has_rel(ID, REL', ID') :
 has_attr(ID', class, CLASS)} = 1 :- choose_rel(TO, TI, CLASS, 
                                                REL, REL', object), 
                                     state(TI, ID).
rel(TO, REL)  :- choose_rel(TO, TI, CLASS, REL, REL', object), state(TI, ID), 
                 has_attr(ID', class, CLASS), has_rel(ID, REL, ID').
rel(TO, REL') :- choose_rel(TO, TI, CLASS, REL, REL', object), state(TI, ID), 
                 has_attr(ID', class, CLASS), has_rel(ID, REL', ID').
bool(TO,yes) :- exist(TO, TI), state(TI,ID).
bool(TO,no)  :- exist(TO, TI), not bool(TO,yes).
bool(TO,no)  :- all_different(TO, TI, ATTR), state(TI, ID), state(TI, ID'), 
                has_attr(ID, ATTR, VALUE), has_attr(ID', ATTR, VALUE).
bool(TO,yes) :- all_different(TO, TI, ATTR), not bool(TO,no).
bool(TO,no)  :- all_same(TO, TI, ATTR), state(TI, ID), state(TI, ID'), 
                has_attr(ID, ATTR, VALUE), not has_attr(ID', ATTR, VALUE).
bool(TO,yes) :- all_same(TO, TI, ATTR), not bool(TO,no).
bool(TO, yes) :- two_different(TO, TI0, TI1, ATTR), state(TI0, ID), 
                 state(TI1, ID'), has_attr(ID, ATTR, VALUE), 
                 has_attr(ID', ATTR, VALUE'), VALUE != VALUE'.
bool(TO, yes) :- two_different(TO, TI0, TI1, ATTR), state(TI0, ID), 
                 state(TI1, ID'), has_attr(ID, ATTR, _), 
                 not has_attr(ID', ATTR, _).
bool(TO, yes) :- two_different(TO, TI0, TI1, ATTR), state(TI0, ID), 
                 state(TI1, ID'), not has_attr(ID, ATTR, _), 
                 has_attr(ID', ATTR, _).
bool(TO,no)   :- two_different(TO, TI0, TI1, ATTR), not bool(TO,yes).
bool(TO, yes) :- two_same(TO, TI0, TI1, ATTR), state(TI0, ID), 
                 state(TI1, ID'), has_attr(ID, ATTR, VALUE), 
                 has_attr(ID', ATTR, VALUE'), VALUE == VALUE'.
bool(TO,no)    :- two_same(TO, TI0, TI1, ATTR), not bool(TO,yes).
attr(TO, ATTR) :- common(TO, TI0, TI1), state(TI0, ID), state(TI1, ID'), 
                  has_attr(ID, ATTR, VALUE), has_attr(ID', ATTR, VALUE), 
                  ATTR != name, ATTR != class, ATTR != hposition, 
                  ATTR != vposition.
{attr(TO, ATTR) : is_attr(ATTR)} = 1 :- common(TO, TI0, TI1).
state(TO,ID) :- compare(TO, TI0, TI1, VALUE, true), state(TI0, ID), 
                state(TI1, ID'), has_attr(ID, _, VALUE), 
                not has_attr(ID', _, VALUE).
state(TO,ID') :- compare(TO, TI0, TI1, VALUE, true), state(TI0, ID), 
                 state(TI1, ID'), not has_attr(ID, _, VALUE), 
                 has_attr(ID', _, VALUE).
state(TO,ID') :- compare(TO, TI0, TI1, VALUE, false), state(TI0, ID), 
                 state(TI1, ID'), has_attr(ID, _, VALUE), 
                 not has_attr(ID', _, VALUE).
state(TO,ID) :- compare(TO, TI0, TI1, VALUE, false), state(TI0, ID), 
                state(TI1, ID'), not has_attr(ID, _, VALUE), 
                has_attr(ID', _, VALUE).
bool(TO,yes) :- and(TO, TI0, TI1), bool(TI0,yes), bool(TI1,yes).
bool(TO,no)  :- and(TO, TI0, TI1), not bool(TO,yes).
bool(TO,yes) :- or(TO, TI0, TI1), bool(TI0,yes).
bool(TO,yes) :- or(TO, TI0, TI1), bool(TI1,yes).
bool(TO,no)  :- or(TO, TI0, TI1), not bool(TO,yes).
{state(TO,ID): state(TI,ID)} = 1 :- unique(TO, TI).
:~ unique(TO, TI), state(TO,ID), has_obj_weight(ID, P). [P, (TO, ID)]
state(TO, ID) :- negate(TO, TI0, TI1), state(TI1, ID), not state(TI0, ID).
is_attr_value(ID, X) :- query(TO, TI, ATTR), state(TI, ID), 
                        has_attr(ID, ATTR, X).
is_attr(X) :- query(TO, TI, ATTR), state(TI, ID), has_attr(ID, ATTR, X).
ans(V) :- end(TO), attr_value(TO,V).
ans(V) :- end(TO), attr(TO,V).
ans(V) :- end(TO), rel(TO,V).
ans(V) :- end(TO), bool(TO,V).
:- not ans(_).
#show ans/1.
\end{Verbatim}
}

\subsection{CLEVR}
\bigskip

\paragraph{CLEVR Question Encoding}

{ \small
\begin{Verbatim}[breaklines=true, breakanywhere=true]
scene(0).
filter_large(1).
filter_green(2).
filter_metal(3).
unique(4).
relate_behind(5).
filter_small(6).
filter_rubber(7).
filter_cube(8).
unique(9).
query_color(10).
scene(11).
filter_large(12).
filter_green(13).
filter_metal(14).
filter_sphere(15).
unique(16).
relate_left(17).
filter_large(18).
filter_rubber(19).
filter_cube(20).
unique(21).
query_color(22).
equal_color(23,11).
end(24).
\end{Verbatim}
}

\paragraph{CLEVR Scene Encoding}

{ \small
\begin{Verbatim}[breaklines=true, breakanywhere=true]
obj(0,129,121,rubber,green,cube,small).
obj(1,131,202,metal,blue,cylinder,small).
obj(2,247,89,rubber,green,cube,small).
obj(3,296,116,metal,red,sphere,large).
obj(4,212,190,metal,yellow,sphere,small).
obj(5,276,158,rubber,yellow,cube,large).
obj(6,70,124,rubber,gray,cube,large).
obj(7,419,149,metal,gray,cylinder,large).
obj(8,330,152,rubber,green,cube,small).
obj(9,190,97,metal,green,sphere,large).
\end{Verbatim}
}

\paragraph{CLEVR ASP Theory}

{ \small
\begin{Verbatim}[breaklines=true, breakanywhere=true]
scene(X) :- scene(X,_).
object(ID) :- obj(ID,_,_,_,_,_,_).
position(ID,X,Y) :- obj(ID,X,Y,_,_,_,_).
has_size(ID,SIZE) :- obj(ID,_,_,_,_,_,SIZE).
has_color(ID,COLOR) :- obj(ID,_,_,_,COLOR,_,_).
has_material(ID,MATERIAL):- obj(ID,_,_,MATERIAL,_,_,_).
has_shape(ID,SHAPE) :- obj(ID,_,_,_,_,SHAPE,_).
left_of(ID,ID')  :- position(ID,X,Y), position(ID',X',Y'), state(T',ID'), 
                    ID!=ID', X<X'.
right_of(ID,ID') :- position(ID,X,Y), position(ID',X',Y'), state(T',ID'), 
                    ID!=ID', X>=X'.
in_front_of(ID,ID') :- position(ID,X,Y), position(ID',X',Y'), state(T',ID'), 
                       ID!=ID', Y>Y'.
behind_of(ID,ID')   :- position(ID,X,Y), position(ID',X',Y'), state(T',ID'), 
                       ID!=ID', Y<=Y'.
state(T+1,ID) :- unique(T), state(T,ID).
:- unique(T), state(T,ID), state(T,ID'), ID!=ID'.
state(T+1,ID) :- relate_left(T), state(T,ID'), left_of(ID,ID').
state(T+1,ID) :- relate_right(T), state(T,ID'), right_of(ID,ID').
state(T+1,ID) :- relate_front(T), state(T,ID'), in_front_of(ID,ID') .
state(T+1,ID) :- relate_behind(T), state(T,ID'), behind_of(ID,ID').
int(T+1,V) :- count(T), #count{ ID : state(T,ID) } = V.
bool(T+1,yes) :- exist(T), state(T,ID).
bool(T+1,no)  :- exist(T), not bool(T+1,yes).
state(T+1,ID) :- filter_large(T), state(T,ID), has_size(ID,large).
state(T+1,ID) :- filter_small(T), state(T,ID), has_size(ID,small).
state(T+1,ID) :- filter_gray(T), state(T,ID), has_color(ID,gray).
state(T+1,ID) :- filter_red(T), state(T,ID), has_color(ID,red).
state(T+1,ID) :- filter_blue(T), state(T,ID), has_color(ID,blue).
state(T+1,ID) :- filter_green(T), state(T,ID), has_color(ID,green).
state(T+1,ID) :- filter_brown(T), state(T,ID), has_color(ID,brown).
state(T+1,ID) :- filter_purple(T), state(T,ID), has_color(ID,purple).
state(T+1,ID) :- filter_cyan(T), state(T,ID), has_color(ID,cyan).
state(T+1,ID) :- filter_yellow(T), state(T,ID), has_color(ID,yellow).
state(T+1,ID) :- filter_metal(T), state(T,ID), has_material(ID,metal).
state(T+1,ID) :- filter_rubber(T), state(T,ID), has_material(ID,rubber).
state(T+1,ID) :- filter_sphere(T), state(T,ID), has_shape(ID,sphere).
state(T+1,ID) :- filter_cylinder(T), state(T,ID), has_shape(ID,cylinder).
state(T+1,ID) :- filter_cube(T), state(T,ID), has_shape(ID,cube).
size(T+1,SIZE) :- query_size(T), state(T,ID), has_size(ID,SIZE).
color(T+1,COLOR) :- query_color(T), state(T,ID), has_color(ID,COLOR).
material(T+1,MATERIAL) :- query_material(T), state(T,ID), 
                          has_material(ID,MATERIAL).
shape(T+1,SHAPE) :- query_shape(T), state(T,ID), has_shape(ID,SHAPE).
state(T+1,ID)  :- and(T,T'), state(T,ID), state(T',ID).
state(T+1,ID)  :- or(T,T'), state(T,ID).
state(T+1,ID') :- or(T,T'), state(T',ID').
bool(T+1, yes) :- boolean_negation(T), bool(T, no).
bool(T+1, no)  :- boolean_negation(T), not bool(T+1, yes).
state(T+1,ID') :- same_size(T), state(T,ID), has_size(ID,SIZE), 
                  has_size(ID',SIZE), ID!=ID'.
state(T+1,ID') :- same_color(T), state(T,ID), has_color(ID,COLOR), 
                  has_color(ID',COLOR), ID!=ID'.
state(T+1,ID') :- same_material(T), state(T,ID), has_material(ID,MATERIAL), 
                  has_material(ID',MATERIAL), ID!=ID'.
state(T+1,ID') :- same_shape(T), state(T,ID), has_shape(ID,SHAPE), 
                  has_shape(ID',SHAPE), ID!=ID'.
bool(T+1,yes) :- equal_integer(T,T'), int(T,V), int(T',V'), V=V'.
bool(T+1,no)  :- equal_integer(T,T'), not bool(T+1,yes).
bool(T+1,yes) :- less_than(T,T'), int(T,V), int(T',V'), V<V'.
bool(T+1,no)  :- less_than(T,T'), not bool(T+1,yes).
bool(T+1,yes) :- greater_than(T,T'), int(T,V), int(T',V'), V>V'.
bool(T+1,no)  :- greater_than(T,T'), not bool(T+1,yes).
bool(T+1,yes) :- equal_size(T,T'), size(T,V), size(T',V'), V=V'.
bool(T+1,no)  :- equal_size(T,T'), not bool(T+1,yes).
bool(T+1,yes) :- equal_color(T,T'), color(T,V), color(T',V'), V=V'.
bool(T+1,no)  :- equal_color(T,T'), not bool(T+1,yes).
bool(T+1,yes) :- equal_material(T,T'), material(T,V), material(T',V'), V=V'.
bool(T+1,no)  :- equal_material(T,T'), not bool(T+1,yes).
bool(T+1,yes) :- equal_shape(T,T'), shape(T,V), shape(T',V'), V=V'.
bool(T+1,no)  :- equal_shape(T,T'), not bool(T+1,yes).
state(0,ID)   :- object(ID).
state(T+1,ID) :- scene(T), object(ID).
ans(V) :- end(T), size(T,V).
ans(V) :- end(T), color(T,V).
ans(V) :- end(T), material(T,V).
ans(V) :- end(T), shape(T,V).
ans(V) :- end(T), bool(T,V).
ans(V) :- end(T), int(T,V).
:- not ans(_).
#show ans/1.
\end{Verbatim}
}

\subsection{\graph}
\bigskip

\paragraph{\graph Question Encoding}

{ \small
\begin{Verbatim}[breaklines=true, breakanywhere=true]
station(0,schwairs).
station(0,pfoiark).
shortestPath(1).
countNodesBetween(2).
end(3).
\end{Verbatim}
}

\paragraph{\graph Scene Encoding}

{ \small
\begin{Verbatim}[breaklines=true, breakanywhere=true]
node(thraonk,orange).
node(mcciourly,orange).
node(pfoiark,orange).
node(shuosh,red).
node(mault,red).
node(mcnaiantly,red).
node(schliws,red).
node(kwaols,red).
node(schwairs,red).
node(hydrioch,red).
node(rhemp,green).
node(whiiny,green).
node(spaarty,brown).
node(yeaums,brown).
node(shehl,brown).
node(typieus,brown).
node(squan,brown).
node(krauess,olive).
node(sairr,olive).
node(greern,olive).
node(skoiaps,olive).
node(boiafy,olive).
node(kruv,olive).
edge(thraonk,mcciourly,orange).
edge(mcciourly,pfoiark,orange).
edge(shuosh,mault,red).
edge(mault,mcnaiantly,red).
edge(mcnaiantly,schliws,red).
edge(schliws,kwaols,red).
edge(kwaols,mcciourly,red).
edge(mcciourly,schwairs,red).
edge(schwairs,hydrioch,red).
edge(pfoiark,rhemp,green).
edge(rhemp,whiiny,green).
edge(spaarty,yeaums,brown).
edge(yeaums,rhemp,brown).
edge(rhemp,shehl,brown).
edge(shehl,typieus,brown).
edge(typieus,squan,brown).
edge(shuosh,krauess,olive).
edge(krauess,sairr,olive).
edge(sairr,greern,olive).
edge(greern,skoiaps,olive).
edge(skoiaps,boiafy,olive).
edge(boiafy,kruv,olive).
line(orange).line(red).
line(green).line(brown).
line(olive).
\end{Verbatim}
}

\paragraph{\graph ASP Theory}

{ \small
\begin{Verbatim}[breaklines=true, breakanywhere=true]
node(S1, gray) :- node(S1, C1), edge(S1,S2,C1), edge(S1,S3,C2), C1 != C2.
edge(S1,S2,C)  :- edge(S2,S1,C), S1!=S2.
node(S1, C)    :- node(S1, gray), edge(S1,_,C).
e(S1,S2) :- edge(S1,S2,C).
e(S2,S1) :- e(S1,S2).
hasNodes(C,N)  :- line(C), node(N,C).
sp((N, N'))  :- shortestPath(T), station(T-1,N), station(T-1,N'), N<N'.
0 { selected((N, N'),S1,S2) ; 
    selected((N, N'),S2,S1) } 1 :- edge(S1,S2,C), sp((N, N')).
path((N, N'),X,Y) :- selected((N, N'),X,Y).
path((N, N'),X,Z) :- path((N, N'),X,Y), path((N, N'),Y,Z).
:- sp((N, N')), not path((N, N'), N, N').
cost((N, N'), C) :- C = #count { X,Y : selected((N, N'),X,Y) }, sp((N, N')).
#minimize { C,N,N' : cost((N, N'), C) }.
hops(N,0) :- withinHops(T, X), station(T-1,N).
hops(N,1) :- not hops(N,0), hops(S,0), e(S,N).
hops(N,2) :- not hops(N,1), hops(S,1), e(S,N).
minHops(N,V) :- V = #min{ X: hops(N,X) }, node(N,_), withinHops(_,_).
countedNodes(T+1, C-1) :- countNodesBetween(T), sp((N, N')), cost((N,N'), C).
p((N,N'))   :- paths(T), station(T-1,N), station(T-1,N'), N<N'.
sp((N,N))   :- cycle(T), station(T-1,N).
cycleAns(T+1,true)  :- cycle(T), path((N,N), N,N).
adjacentAns(T+1, true)  :- adjacent(T), station(T-1,N), 
                           station(T-1,N'), N<N', e(N,N').
adjacentAns(T+1, false) :- adjacent(T), not adjacentAns(T+1, true).
adjacentToAns(T+1,X)    :- adjacentTo(T), station(T-1,N), station(T-1,N'), 
                           N<N', e(N,X), e(N',X).
start(S1)   :- p((S1,S2)).
finish(S2)  :- p((S1,S2)).
{ in(X,Y) : e(X,Y) }.
:- in(X,Y), in(Z,Y), X != Z.
:- in(X,Y), in(X,Z), Y != Z.
:- start(Y), in(_,Y).
:- finish(X), in(X,_).
reach(X) :- start(X), in(X,_).
reach(Y) :- in(X,Y), reach(X).
:- finish(X), not reach(X).
:- in(X,Y), not reach(X).
:- in(X,Y), not reach(Y).
sameLineAns(T+1,true)   :-  sameLine(T), station(T-1,N), station(T-1,N'), 
                            hasNodes(L,N), hasNodes(L,N'), N<N'.
sameLineAns(T+1,false)  :-  sameLine(T), not sameLineAns(T+1,true).
sp((N,N'))  :- commonStation(T), station(T-1,N), station(T-1,N'), N<N'.
commonStationAns(T+1, true)  :- commonStation(T), cost((N,N'), C), C = 2.
commonStationAns(T+1, false) :- commonStation(T), 
                                not commonStationAns(T+1, true).
lineOnAns(T+1,LN)   :- linesOnNames(T), station(T-1,N), hasNodes(LN,N).
lineOnAns(T+1,C)    :- linesOnCount(T), station(T-1,N), 
                       C=#count{L : hasNodes(L,N), L!=gray}.
existAns(T+1,true)  :- exist(T), station(T-1,N), node(N,C).
existAns(T+1,false) :- exist(T), not existAns(T+1,true).
stationsAns(T+1, N) :- stations(T), line(T-1,LN), hasNodes(LN,N).

ans(V) :- end(T), countedNodes(T,V).
ans(V) :- end(T), cycleAns(T,V).
ans(V) :- end(T), adjacentAns(T,V).
ans(V) :- end(T), adjacentToAns(T,V).
ans(V) :- end(T), sameLineAns(T,V).
ans(V) :- end(T), commonStationAns(T,V).
ans(V) :- end(T), lineOnAns(T,V).
ans(V) :- end(T), existAns(T,V).
ans(V) :- end(T), stationsAns(T,V).
:- not ans(_).
#show ans/1.
\end{Verbatim}
}

\section{Prompts}
\label{apdx:prompts}
This section contains the full prompts used in our rule distillation process, mentioned in Section~\ref{sec:distillation}. 
We show the system preprompts and examples of how different prompting strategies were applied, including multi-prompting, chain-of-thought reasoning, and syntactic/semantic mending.
\subsection{GQA}
\bigskip

\paragraph{GQA Scene Representation}
{\small
\begin{Verbatim}[breaklines=true, breakanywhere=true]
Consider the following ASP representation for the objects in an image.

Object Declaration: Each object(<id>) declares a
unique object with a specific identifier. 
For example, object(1279158) declares an object with the ID 1279158.
Attributes (has_attr): The 
has_attr(<object_id>, <attribute>, <value>) facts define various attributes 
of an object. 
For example, has_attr(1279158, class, tree) means the object with ID 1279158 is
classified as a "tree". Attributes cover a range of properties like class, name, 
color, hposition (horizontal position), vposition (vertical position), and more.
Relations (has_rel): The has_rel(<object_id1>, <relation>, <object_id2>) facts 
describe the relationships between two objects. 
For example, has_rel(1279158, to_the_right_of, 1279150) indicates that the 
object 1279158 is to the right of the object 1279150.
Each object in this set of facts seems to represent an element in a larger scene 
or model, 
with attributes that describe its characteristics (like class, color, position) 
and relations that describe its spatial or logical connections to other elements 
in the scene. 
For example, objects are classified into categories like tree, wheel, wing, 
etc., and their positions are described in relation to other objects (e.g., 
to_the_right_of, below, near).

An excerpt of an example of such an encoding is the following:

object(1120957).
has_attr(1120957, class, jet).
has_attr(1120957, vposition, middle).
object(1120970).
has_attr(1120970, class, window).
has_rel(1120970, to_the_right_of, 1120977).
has_rel(1120970, to_the_right_of, 1120980).
object(1120964).
has_attr(1120964, class, shirt).
has_attr(1120964, name, shirt).
has_attr(1120964, vposition, middle).
has_rel(1120964, to_the_left_of, 1120963).
...
\end{Verbatim}
}
\paragraph{GQA Question Representation}
{\small
\begin{Verbatim}[breaklines=true, breakanywhere=true]
Consider the following ASP representation for natural language questions.
The representation shows the steps needed to solve the question.
The first number of each predicate indicates the output step.
The rest of the numbers indicate the input steps.
In this way chains of predicates are joint together to represent the
reasoning steps needed to answer the question.

scene(S): Initializes all objects at step S.
select(S, I, O): Selects object O at step S, based on input I.
relate(S, I, O, R, Sub): Establishes relationship R between O and Sub at S.
unique(S, I): Asserts uniqueness of an object from I at S.
query(S, I, A): Queries attribute A of an object from I at S.
end(S): Concludes query at S.
filter(S, I, A, V): Filters objects by A and V at S.
relate_any(S, I, R, Sub): Establishes a general R at S.
filter_any(S, I, A): Filters objects by A presence at S.
negate(S, I, P): Negates condition from I at S.
exist(S, I): Checks existence of I at S.
verify_attr(S, I, A, V): Verifies A with V of I at S.
all_same(S, I, A): Checks if all from I share A.
choose_attr(S, I, A, O1, O2): Chooses between O1 and O2 for A at S.
choose_rel(S, I, O, R1, R2, Sub): Chooses between R1 and R2 at S.
common(S, I1, I2): Finds commonality between I1 and I2 at S.
two_different(S, I1, I2, A): Compares I1 and I2 for different A at S.
two_same(S, I1, I2, A): Compares I1 and I2 for same A at S.
and(S, I1, I2): Logical AND of I1 and I2 at S.
or(S, I1, I2): Logical OR of I1 and I2 at S.
compare(S, I1, I2, C, B): Compares I1 and I2 based on condition C and boolean B at S.

Examples of questions are the following, separated by rows of #:
scene(0).
select(1, 0, animal).
all_same(2, 1, class).
end(2).
#########
scene(0).
all_different(2, 1, class).
end(2).
#########
scene(0).
select(1, 0, pizza).
exist(2, 1).
end(2).
\end{Verbatim}
}

\paragraph{CLEVR Scene Representation}
{\small
\begin{Verbatim}[breaklines=true, breakanywhere=true]
Consider the following ASP representation for the objects in an image.

It consists of possibly multiple predicates of the form 'obj(ID,X,Y,M,C,F,S)',
where 'ID' is a unique ID that defines the object, 'X, Y' are coordinates 
between 0 and 10, 
'M' is a material, 'C' is a color, 'F' is a form and 'S' is a size.
An excerpt of an example of such an encoding is the following:

obj(0,324,201,rubber,purple,sphere,large).
obj(1,282,166,rubber,purple,cylinder,small).
obj(2,216,94,metal,blue,sphere,large).
obj(3,127,115,metal,green,cube,large).
...
\end{Verbatim}
}

\subsection{CLEVR}
\bigskip

\paragraph{CLEVR Question Representation}
{\small
\begin{Verbatim}[breaklines=true, breakanywhere=true]
Consider the following ASP representation for natural language questions.
The representation shows the steps needed to solve the question.
The first number of each predicate indicates the output step.
The rest of the numbers indicate the input steps.
In this way chains of predicates are joint together to represent the reasoning 
steps needed to answer the question.

scene(X): 
    Indicates the start of a new scene or a situation.
    X: Identifier for the scene.

filter_property(X)
    Applies a filter based on a property (e.g., color, size, material) 
    to objects in the scene.
    X: Identifier for the filter operation, or a specific property value.

unique(X)
    Specifies that the previous filter operation results in a unique object.
    X: Identifier linked to the outcome of the filter.

relate_direction(X):
    Defines a spatial relationship between objects.
    X: Specifies the identifier for the relationship.

and(X, Y)
    Logical AND operation, often used to combine conditions.
    X, Y: Identifiers of the conditions or scenes to be combined.

and(X, Y)
    Logical OR operation, often used to combine conditions.
    X, Y: Identifiers of the conditions or scenes to be combined.

query_attribute(X)
    Requests information about a particular attribute 
    (e.g., color, material, shape) of the object.
    X: Identifier for the query operation.

exist(X)
    Asserts the existence of objects meeting previous criteria.
    X: Optionally identifies the particular existence check.

end(X)
    Marks the end of the query or scene description.
    X: Identifier that concludes the scene or query.

equal_[attribute](X, Y)
    Compares two attributes to assert equality.
    X, Y: Identifiers of the attributes or scenes being compared.

same_attribute
    Asserts that two or more objects share the same attribute 
    (e.g., size, material).
    X: Identifier for the comparison operation.

count(X)
    Count the objects that meet the criteria set before this predicate.
    X: Identifier for the count operation.

Examples of questions are the following, separated by rows of #:
scene(0).
filter_large(1).
filter_green(2).
filter_metal(3).
unique(4).
same_shape(5).
filter_small(6).
filter_yellow(7).
exist(8).
end(9).
##############
scene(0).
filter_blue(1).
unique(2).
relate_right(3).
filter_rubber(4).
filter_cylinder(5).
count(6).
scene(7).
filter_large(8).
filter_rubber(9).
filter_cube(10).
unique(11).
relate_behind(12).
filter_brown(13).
filter_rubber(14).
filter_cylinder(15).
count(16).
equal_integer(17,7).
end(18).
##############       
\end{Verbatim}
}

\subsection{\graph}
\bigskip

\paragraph{\graph Scene Representation}
{\small
\begin{Verbatim}[breaklines=true, breakanywhere=true]
Consider the following ASP representation for a metro graph.

It consists of possibly multiple predicates of the form 'node(N,C)', where N is
the name of the station and C is the color,
'edge(N1,N2,C)', where N1,N2 are names and C is the color of the edge and 
'line(C)', where C is the color of the metro line.

An excerpt of an example of such an encoding is the following:

node(vuirst,olive).
node(schnoms,olive).
node(mccreofy,olive).
node(crirg,olive).
edge(crirg,schnoms,cyan).
edge(reiards,niath,pink).
edge(niath,skoack,pink).
line(olive).
line(cyan).
line(pink).
.....    
\end{Verbatim}
}
\paragraph{\graph Question Representation}
{\small
\begin{Verbatim}[breaklines=true, breakanywhere=true]
Consider the following ASP representation for natural language questions.
The representation shows the steps needed to solve the question.

The representation is related to a problem involving graph traversal or analysis, 
where nodes represent stations and edges represent connections (or lines) 
between those stations. 
The variables and predicates in the questions can be interpreted as follows:

end(stepNumber):

    Description: Indicates the end of the computation or query process. The 
    stepNumber defines the step number at which this stage occurs.
    Variables: stepNumber (an integer indicating the computation step).

countNodesBetween(stepNumber):

    Description: Counts the number of nodes between two specified stations 
    or nodes.
    Variables: stepNumber (an integer indicating the computation step).

shortestPath(stepNumber):

    Description: Finds the shortest path between two stations or nodes.
    Variables: stepNumber (an integer indicating the computation step).

station(stepNumber, stationName):

    Description: Refers to a specific station in the context of the query. 
    The stationName can be replaced with the name or identifier of a station.
    Variables: stepNumber (an integer indicating the computation step), 
    stationName (a string or variable representing the station's name).

paths(stepNumber):

    Description: Identifies paths between stations or nodes
    Variables: stepNumber (an integer indicating the computation step).

cycle(stepNumber):

    Description: Identifies a cycle 
    (a path that starts and ends at the same station or node).
    Variables: stepNumber (an integer indicating the computation step).

adjacent(stepNumber):

    Description: Identifies nodes or stations that are adjacent to each other.
    Variables: stepNumber (an integer indicating the computation step).

adjacentTo(stepNumber):

    Description: Finds stations or nodes adjacent to a specific station or node.
    Variables: stepNumber (an integer indicating the computation step).

adjacentArch(stepNumber, archName):

    Description: Finds arches adjacent to a specified station or node.
    Variables: stepNumber (an integer indicating the computation step), 
    archName (a string or variable representing the arch's name).

commonStation(stepNumber):

    Description: Finds common stations between paths or connections.
    Variables: stepNumber (an integer indicating the computation step).

exist(stepNumber):

    Description: Checks for the existence of a specific station or node 
    in the network.
    Variables: stepNumber (an integer indicating the computation step).

linesOnNames(stepNumber):

    Description: Identifies lines based on their names.
    Variables: stepNumber (an integer indicating the computation step).

linesOnCount(stepNumber):

    Description: Counts the number of lines passing through a station or node.
    Variables: stepNumber (an integer indicating the computation step).

sameLine(stepNumber):

    Description: Checks whether two or more stations or nodes are 
    on the same line.
    Variables: stepNumber (an integer indicating the computation step).

stations(stepNumber):

    Description: Refers to multiple stations, possibly in relation 
    to a line or path.
    Variables: stepNumber (an integer indicating the computation step).

line(stepNumber, lineName):

    Description: Refers to a specific line in the context of the query. 
    The lineName can be replaced with the name or identifier of a line.
    Variables: stepNumber (an integer indicating the computation step),
    lineName (a string or variable representing the line's name).

Examples of questions are the following, separated by rows of #:
end(2).
linesOnNames(1).
station(0, joith).
##########
end(2).
commonStation(1).
station(0, pfarm).
station(0, khieuty).
##########
end(2).
cycle(1).
station(0, striaolt).
\end{Verbatim}
}

\subsection{Uniform (dataset-independent) subprompts}

\medskip
\paragraph{Preprompt}

{ \small
\begin{Verbatim}[breaklines=true, breakanywhere=true]
1. Contextualisation
We are in the domain of Visual Question Answering.
The task consists of taking an image and question related to it as input 
and produce as output the correct answer.

We have already preprocessed both the image and question into correct 
Answer Set Programming representations.
Scene/Question pairs of ASP facts serve as the instance for an ASP program 
which we call the Theory.
This is a collection of rules that handles the input instance 
and calculates the correct answer.

Your task is to help us expand the Theory with new rules as new instances 
of questions appear.
We will first give you some definitions in the following sections 
and then present the concrete task.

2. Answer Set Programming Syntax

Answer Set Programming (ASP) is a form of declarative programming 
oriented towards difficult search problems. 
Its syntax and usage can be summarized as follows:

Rules: The basic building block of an ASP program. A rule has a head and a body,
and is written in the form: Head :- Body. 
It means that if the body is true, then the head is also true.
Predicates in the Body are separated by commas, and not semicolons as in Prolog.
For example: flies(tweety) :- bird(tweety), not penguin(tweety).

Atoms: These are the basic propositions and can be any string of characters 
and numbers starting with a lowercase letter.

Literals: An atom or its negation. The negation used in ASP is 
negation as failure, denoted by not. 
For example, not a means that a cannot be proven to be true.

Facts: These are rules without a body, stating something 
that is unconditionally true. 
For example: bird(tweety).

Constraints: These are rules without a head, used to eliminate certain answers. 
For example, :- not fly(tweety). states that any answer set 
where tweety does not fly is not acceptable.

Choice Rules: These allow for the generation of multiple answers, 
expressing that atoms can be freely chosen to be in the answer set. 
For example, {fish(tweety);bird(tweety)} :- penguin(tweety). 
means that the penguin tweety may be a fish, a bird or none.

Comments: In ASP, comments start with a % and continue to the end of the line.
They are ignored by the ASP solver.

In ASP, the use of ; to represent logical disjunction in the body 
of a rule is 
not allowed. 
Instead, we need to express disjunction through separate rules. 

In ASP, a variable in a rule is considered unsafe when a variable 
appears in the head of a rule or in a negative literal in the body, 
but not in a positive literal.
Examples of Unsafe Variables:

% Example 1: Unbound Y in negated predicate, S and T in regular predicates
p1(X) :- not q(X,Y), r(S), s(T).

% Example 2: Unbound Y in negated predicate, S in regular predicate, T in negated predicate
p2(X, B) :- not q(Y), r(S), not s(T).

3. Scene Representation

{scene_representation}

4. Question Representation

{question_representation}

6. Initial ASP Theory

{initial_asp_theory}

7. Task.

Your task is to extend the initial ASP theory we provided with new rules 
that allows us to calculate the correct answer.
You will be presented with a VQA example containing an scene, question 
and expected answer.
Return ASP rules that are able to correctly process the example.

THIS IS VERY IMPORTANT! 
IF YOU BREAK ANY OF THESE RULES THE OUTPUT IS WRONG!

1. Only output the new ASP rules.
2. Do not output ASP facts.
3. New rules should be as general as possible, i.e., have a low number 
of constants and high number of variables.
4. Do not use Code Block Format when returning the new rules, just plain text.
5. Do not add any unsafe rules.

\end{Verbatim}
}


\paragraph{Multi-Prompting}

{ \small
\begin{Verbatim}[breaklines=true, breakanywhere=true]
Consider you must output N different possibilities of the rules 
that need to be implemented.
The rule sets must be separated by a line of three asterisks.
\end{Verbatim}
}
\medskip

\paragraph{Simple Chain-of-Thought}

{ \small
\begin{Verbatim}[breaklines=true, breakanywhere=true]
You are also required to produce a chain of thought using the predicate
in the theory,
showing how you arrived to the correct answer using them.
Output a line for each predicate used.
Separate the response rules from the chain of thought by a line of 
three hashtag symbols.
The first predicate must be scene and the last predicate must be ans.
Only output one line per predicate used. 
Do not output state predicates.
No rules should be below the three asterisks.
\end{Verbatim}
}
\medskip

\paragraph{Advanced Chain-of-Thought}

{ \small
\begin{Verbatim}[breaklines=true, breakanywhere=true]
You are also required to produce a chain of thought using the predicates in the theory,
showing how you arrived to the correct answer using them.
Follow the steps below to break down domain knowledge and construct the rules:

1. **Understand the Problem:**
   Briefly describe the problem domain and the specific goal.

2. **Break Down the Domain Knowledge:**
   Identify key concepts and entities in the problem domain.
   List relevant properties and relationships between these entities.

3. **Formulate If-Then Statements:**
   For each identified relationship or property, write an if-then statement that captures the logic of the relationship.
   Ensure that each statement is clear and logically sound.

4. **Construct Logic Rules:**
   Convert the if-then statements into clingo syntax.
   Ensure that the rules are syntactically correct and cover the intended logic comprehensively.

**Example:**

Problem Description:
We need to create a schedule for classes in a school. Classes should not overlap, each teacher can teach only one class at a time, and classrooms can host only one class at a time.

Step-by-Step Process:
1. **Understand the Problem:**
   Goal: Create a schedule without conflicts in class timings, teacher assignments, and classroom usage.

2. **Break Down the Domain Knowledge:**
   Key concepts: Classes, Teachers, Classrooms, Timeslots.
   Relationships: A class has a teacher, a class is held in a classroom, a class occupies a timeslot.

3. **Formulate If-Then Statements:**
   If a class is assigned to a timeslot, then no other class can be in the same classroom at that timeslot.
   If a class is assigned to a teacher, then that teacher cannot teach another class at the same timeslot.
   If a class is scheduled in a classroom, then that classroom cannot be used by another class at the same timeslot.

4. **Construct Logic Rules:**
   % Facts about classes, teachers, classrooms, and timeslots
   class(c1). class(c2).
   teacher(t1). teacher(t2).
   classroom(r1). classroom(r2).
   timeslot(1). timeslot(2).

   % Assigning a class to a timeslot, a teacher, and a classroom
   1 { assign(C, T, R, TS) : teacher(T), classroom(R), timeslot(TS) } 1 :- class(C).

   % Ensure no class overlaps in the same classroom at the same timeslot
   :- assign(C1, T1, R, TS), assign(C2, T2, R, TS), C1 != C2.

   % Ensure no teacher is teaching more than one class at the same timeslot
   :- assign(C1, T, R1, TS), assign(C2, T, R2, TS), C1 != C2.

   % Ensure a classroom hosts only one class at a time
   :- assign(C1, T1, R, TS), assign(C2, T2, R, TS), C1 != C2.

You must first Output the new rules.
Then a line with three # symbols.
Last, the reasoning from the chain of thought.

\end{Verbatim}
}

\medskip

\paragraph{Syntax Mending}

{ \small
\begin{Verbatim}[breaklines=true, breakanywhere=true]
You must repair the syntax of the prompted Answer Set Programming rule(s). 
Additionally, clingo outputed the following error: {syntax_error}.
You must only output the fixed ASP rule(s) and any other rules included 
in the prompt that have correct syntax. Do not output any natural language. 
The output must be in plain text only! 
Do not output the response as a code block!
\end{Verbatim}
}

\medskip

\paragraph{Semantic Mending}

{ \small
\begin{Verbatim}[breaklines=true, breakanywhere=true]
You must repair the semantics of the prompted Answer Set Programming rule(s).
The original theory is: {theory}. 
The prompted rules are added to this theory to calculate the answer. 
They do not calculate the correct answer, which is: {expected_answer}, 
but instead result in the following incorrect answer: {infered_answer} 
(can me empty).
You must only output the fixed ASP rule(s). 
The rule(s) must be semantically different from the ones prompted 
and they should now calculate the correct answer.
Do not output any natural language. 
The output must be in plain text only! 
Do not output the response as a code block!
\end{Verbatim}
}

\section{Ablation Experiments Tables}\label{apdx:ablation}
\medskip
Here, we provide a detailed breakdown of ablation results per predicate.
For each model and prompting strategy, we report the accuracy achieved on individual predicates, allowing a deeper insight on what types of reasoning benefit most from each strategy.

\begin{figure*}[!htbp]
    \centering
    \begin{subfigure}{0.925\textwidth}
        \centering
        \includegraphics[width=\textwidth]{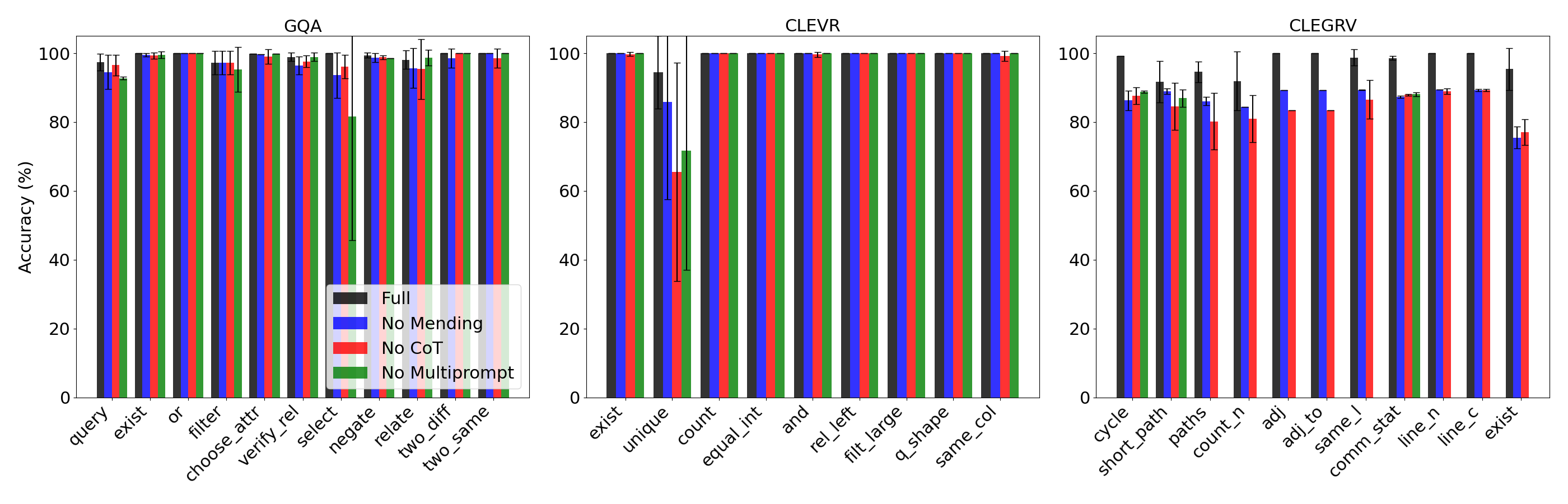}
        \vspace{-2.5em}
        \caption{\footnotesize GPT-4o}
        \label{fig:gpt4o_all_datasets}
    \end{subfigure}
    \begin{subfigure}{0.925\textwidth}
        \centering
        \includegraphics[width=\textwidth]{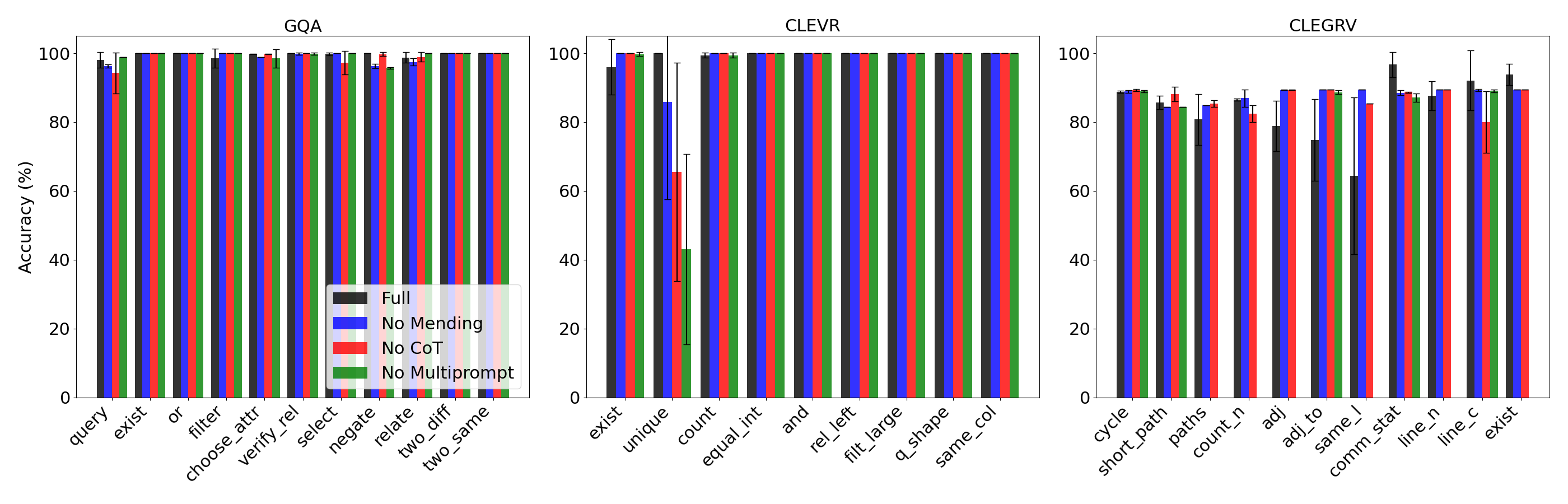}
        \vspace{-2.5em}
        \caption{\footnotesize  DeepSeek}
        \label{fig:deepseek_all_datasets}
    \end{subfigure}
    \begin{subfigure}{0.925\textwidth}
        \centering
        \includegraphics[width=\textwidth]{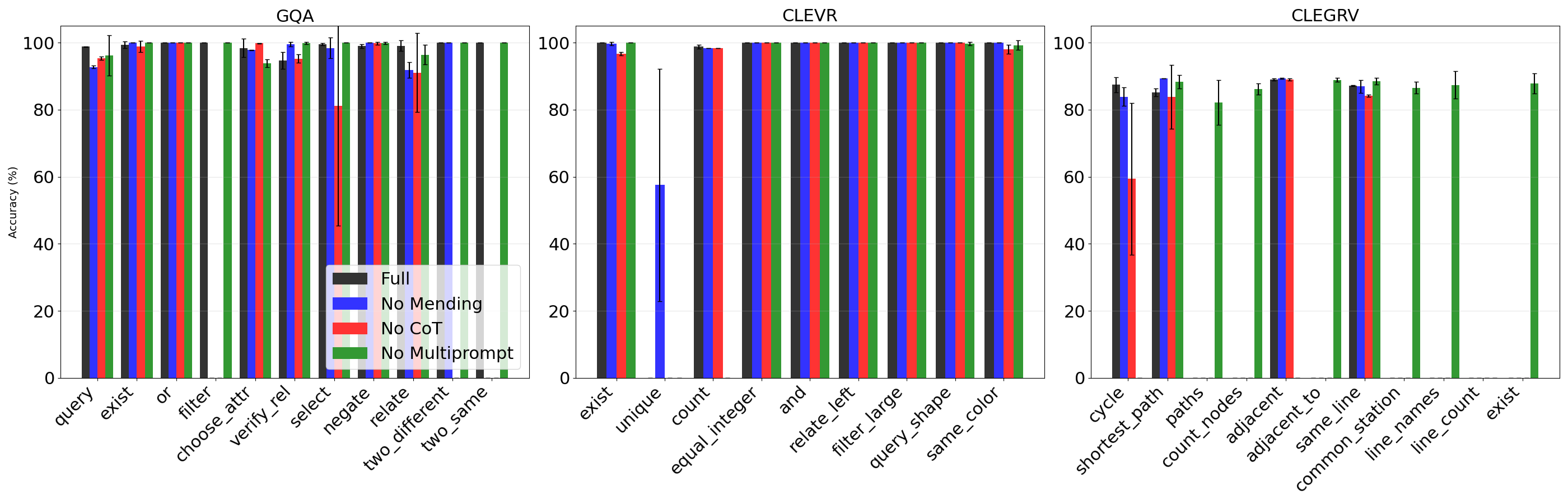}
        \vspace{-2.5em}
        \caption{\footnotesize Mistral-Large}
        \label{fig:mistral_all_datasets}
    \end{subfigure}
    \begin{subfigure}{0.925\textwidth}
        \centering
        \includegraphics[width=\textwidth]{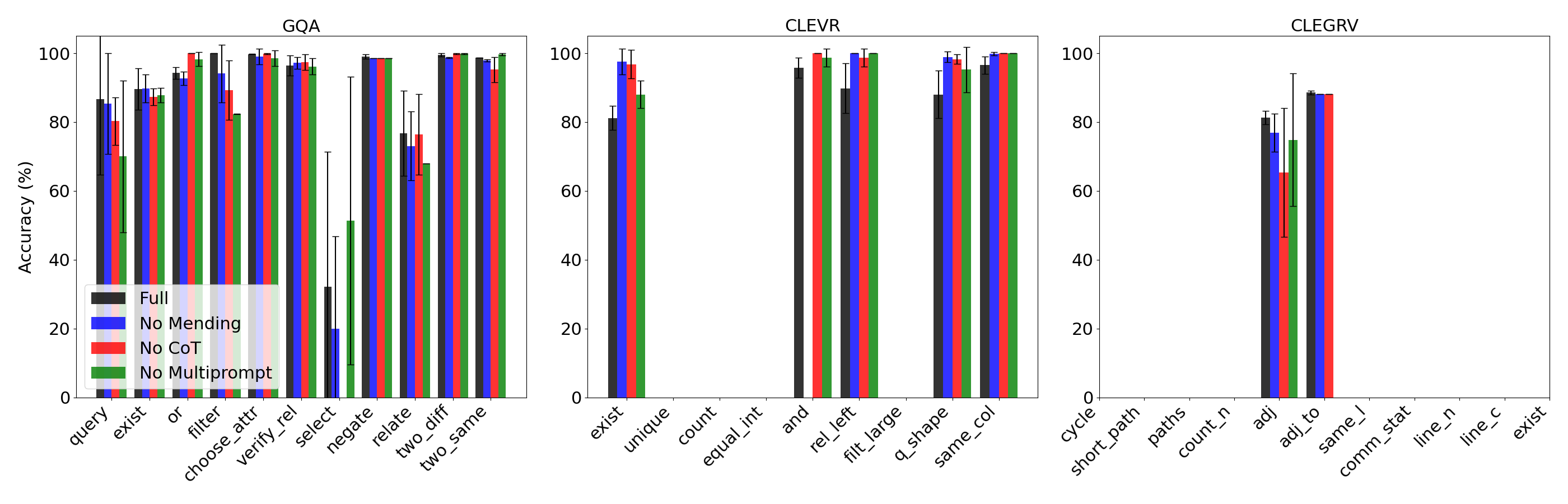}
        \vspace{-2.5em}
        \caption{\footnotesize  LLama3-70b}
        \label{fig:llama_all_datasets}
    \end{subfigure}
    \begin{subfigure}{0.925\textwidth}
        \centering
        \includegraphics[width=\textwidth]{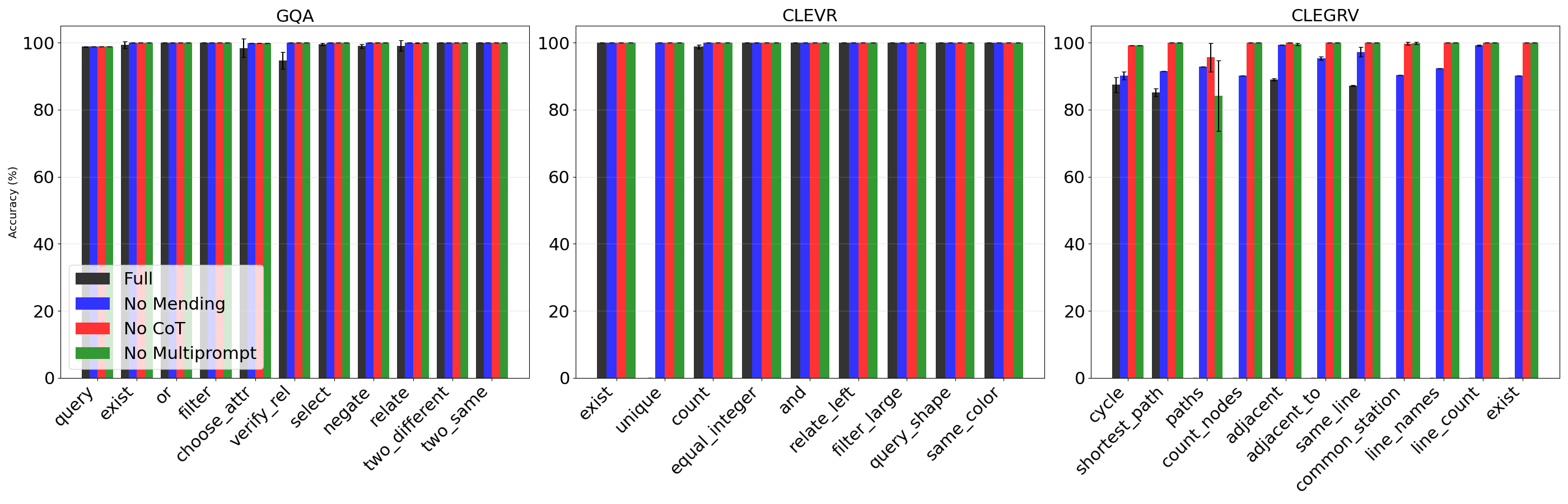}
        \vspace{-2.5em}
        \caption{\footnotesize Gemini-3}
        \label{fig:gemini_all_datasets}
    \end{subfigure}
    \caption{\small
    Comparison of model performance across GQA, CLEVR, and \graph datasets for the experiment without ablation, no mending, no CoT settings and no multiprompt strategies.}
    \label{fig:all_models_comparison}
\end{figure*}

We present the complete tables for each dataset and each ablation mode; Tables~\ref{tab:ablation_mending_GQA}, \ref{tab:ablation_cot_GQA} and~\ref{tab:ablation_multiprompt_GQA} show the results for GQA, Tables~\ref{tab:ablation_mending_CLEVR}, \ref{tab:ablation_cot_CLEVR} and~\ref{tab:ablation_multiprompt_CLEVR} show the results for CLEVR and Tables~\ref{tab:ablation_mending_CLEGR}, \ref{tab:ablation_cot_CLEGR} and~\ref{tab:ablation_multiprompt_CLEGR} for \graph. 

The tables are visualized in Fig.~\ref{fig:all_models_comparison}, which contains four subfigures, each summarizing the ablation results for a different model across the GQA, CLEVR, and \graph{} datasets. Each subfigure reports accuracy averaged over five runs, along with the corresponding standard deviation as error bars, illustrating the impact of removing a distillation strategy.

Subfig.~\ref{fig:gpt4o_all_datasets} presents results for GPT-4o, where our main distillation method surpasses most of the ablation studies as expected.
On GQA, ablation studies are $\approx 2\%$ lower in accuracy than the main method, with a notable exception for the \texttt{select} predicate, where the absence of multiprompt reduced the accuracy by 19\%. On CLEVR, the distillation strategies did not seem to impact most of the predicates; the only predicate noticeably affected was \texttt{unique}, where the average accuracy was reduced from 94\% to $\approx 70\%$, with CoT being the most impactful strategy. On \graph, the distillation strategies are most noticeable for the model, where all predicates appear affected with $\approx 15\%$. Of particular interest are a set of predicates (\texttt{paths}, \texttt{count nodes}, etc.) that were not able to be properly distilled when the multiprompt strategy is deactivated.

Subfig.~\ref{fig:deepseek_all_datasets} shows the results for DeepSeek, which in general performed similarly to GPT-4o with some exceptions. First, some results indicate that removing some of the distilling strategies does not decrease accuracy, but instead increases it. For example, in GQA the predicate \texttt{query} performs worst when using multi-prompting in this model, same for \texttt{exist} in CLEVR or \texttt{adjacent to} in \graph when using any distillation strategy. A potential explanation is the fact that some models may react negatively to multiple prompting stages, where the effects of one stage may influence the following ones. Overburdening systems with prompts may limit their reasoning capabilities, especially on less powerful models. 
We can also observe the same behaviour for \texttt{unique} in CLEVR as for GPT-4o, where all strategies seem effective, but in this case multi-prompting is better than the rest.
Also for \graph, the deactivation of multi-prompting seems to be remarkably impactful in a similar set of predicates as the one for GPT-4o, as no set of rules was produced with the configuration.

Subfig.~\ref{fig:mistral_all_datasets} shows the performance of Mistral-Large. For GQA, the distillation strategies show mixed results. The main method was most effective for the predicates \texttt{query}, \texttt{select}, \texttt{relate}, where deactivating the mending, Cot and CoT respectively decreased the accuracy by approximately 8\%, 20\% and 7\%. 
The effect of these two strategies is more pronounced on the \texttt{filter} predicate, where no rules where extracted.
On the other hand, deactivating some distillation strategies yielded improved results, \egc for \texttt{verify\_rel}, where all ablation studies were superior, up to a margin of 5\% for mending and multi-prompting. 
The CLEVR dataset shows much better results in general but the effect of the strategies is not very visible, with the exception of deactivating CoT for \texttt{exists}, which reduces accuracy by 5\%, and removing multi-prompting for \texttt{count} improves accuracy by 3\%. For \texttt{unique}, no rules were not produced on any configuration. 
\graph showed a mixed reaction to the strategies: deactivating multi-prompting did not produce rules, whereas for \texttt{shortest\_path} and \texttt{common\_station}, the multiprompt study achieved improvements of 4\% and 1\% respectively. Many predicates where not possible to distill using this model, such as \texttt{paths}, \texttt{count nodes} and \texttt{line count}, independent of the configuration. 

Subfigure~\ref{fig:llama_all_datasets} shows the results for LLaMA3-70b. GQA already shows a very increased variance for across all configurations. The ablation studies are mixed; for predicates such as \texttt{query}, \texttt{exist} and \texttt{filter}, deactivating CoT or multipromt led to worse results. The effect of CoT is more strong on the \texttt{select} predicate where no answer was possible to get, where as other configurations did produce rules. In particular, for this predicate multiprompt which achieved an increase of $\approx 19\%$. Nevertheless, the standard deviation is particularly high, which may indicate the results are not consistent. We also see that removing CoT when distilling the \texttt{or} predicate results in a perfect score with zero variance.
For CLEVR, deactivating any of the distillation strategies resulted in improved accuracy. This can be attributed to the performance of the model, which is overall the worst, which indicates that it may be overburdened by the strategies. The most impactful strategy to deactivate is mending, where we see increases for the predicates \texttt{exist}, \texttt{relate left}, \texttt{query shape} and \texttt{same color}, with improvements of 16\%, 11\%, 10\% and 3\% respectively. When deactivating the strategy on the \texttt{and} predicate, no rules were produced while others configuration did.
The results on \graph show that the distillation capabilities of the model are below average no rules were, independent of the configuration.
Most of the predicates could not be distilled with the exception of two, \texttt{adjacent} and \texttt{adjacent to}, in the former deactivating CoT decreased the accuracy by 16\%, while in the latter deactivating multiprompt yielded no rules.

Finally, Subfigure~\ref{fig:gemini_all_datasets} details the performance of Gemini-3. This model exhibits the most robust performance across all datasets, demonstrating a significant ceiling effect. On both GQA and CLEVR, Gemini-3 achieves nearly perfect accuracy ($\approx 100\%$) with negligible variance across all ablation studies, indicating that the removal of mending, CoT, or multi-prompting has minimal impact on its reasoning capabilities for these tasks.
For \graph, the results are similarly high but reveal interesting nuances. Unlike other models where removing strategies often degrades performance, Gemini-3 achieves its best results in \graph{} when Chain-of-Thought is deactivated, reaching $100\%$ accuracy on complex predicates such as \texttt{shortest\_path} and \texttt{adjacent}. In contrast, the absence of mending leads to a slight performance drop (averaging $\approx 90\%$ accuracy) and increased variance. This suggests that while Gemini-3 benefits from mending to stabilize its outputs, the intermediate reasoning steps of CoT may be unnecessary or even slightly restrictive for this model on graph traversal tasks, as it is capable of deriving the correct rules directly.

\begin{table}[tb]
\centering
\caption{GQA Performance without mending.}
\begin{tabular}{lrrrrr}
\toprule
$P$ & GPT-4o & DeepSeek & Mistral-Large & LLama3-70b & Gemini-3\\
\midrule
query           & $94.51 \pm 4.94$ & $96.29 \pm 0.50$ & $92.76 \pm 0.42$  & $85.32 \pm 14.65$  & $98.92 \pm 0.00$  \\
exist           & $99.53 \pm 0.56$ & $99.98 \pm 0.01$ & $99.99 \pm 0.01$ & $89.72 \pm 4.03$  & $100.0 \pm 0.00 $\\
or              & $100.0 \pm 0.00$ & $100.0 \pm 0.00$ & $100.0 \pm 0.00$ & $92.67 \pm 1.91$  & $100.0 \pm 0.00 $\\
filter          & $97.24 \pm 3.38$ & $100.0 \pm 0.00$ & -- & $94.08 \pm 8.37$  & $100.0 \pm 0.00 $\\
choose\_attr    & $99.63 \pm 0.00$ & $98.83 \pm 0.00$ & $97.80 \pm 0.05$ & $99.06 \pm 2.30$  & $99.83 \pm 0.00$\\
verify\_rel     & $96.40 \pm 2.59$ & $99.84 \pm 0.31$ & $99.60 \pm 0.69$ & $97.17 \pm 1.70$  & $100.0 \pm 0.00 $\\
select          & $93.64 \pm 6.58$ & $100.0 \pm 0.00$ & $98.44 \pm 3.12$ & $19.99 \pm 26.82$  & $100.0 \pm 0.00 $\\
negate          & $98.69 \pm 1.34$ & $96.28 \pm 0.72$ & $100.0 \pm 0.00$ & $98.56 \pm 0.00$ & $100.0 \pm 0.00 $ \\
relate          & $95.69 \pm 5.82$ & $97.48 \pm 1.04$ & $91.83 \pm 2.34$ & $73.04 \pm 10.02$  & $100.0 \pm 0.00 $\\
two\_different  & $98.62 \pm 2.76$ & $100.0 \pm 0.00$ & $100.0 \pm 0.00$ & $98.71 \pm 0.21$  & $100.0 \pm 0.00 $\\
two\_same       & $100.0 \pm 0.00$ & $100.0 \pm 0.00$ & -- & $97.83 \pm 0.31$  & $100.0 \pm 0.00 $\\
\bottomrule
\end{tabular}
\label{tab:ablation_mending_GQA}
\end{table}

\begin{table}[tb]
\centering
\caption{CLEVR results without mending.}
\begin{tabular}{lrrrrr}
\toprule
$P$ & GPT-4o & DeepSeek & Mistral-Large & LLama3-70b  & Gemini-3\\
\midrule
exist           & $100.0 \pm 0.00$ & $100.0 \pm 0.00$ & $99.72 \pm 0.56$ & $97.56 \pm 3.75$  & $100.0 \pm 0.00 $\\
unique          & $85.84 \pm 28.32$& $100.0 \pm 0.00$ & $57.52 \pm 34.68$& --  & $100.0 \pm 0.00 $\\
count           & $100.0 \pm 0.00$ & $99.12 \pm 0.73$ & $98.41 \pm 0.00$ & --   & $100.0 \pm 0.00 $\\
equal\_integer  & $100.0 \pm 0.00$ & $100.0 \pm 0.00$ & $100.0 \pm 0.00$ & --  & $100.0 \pm 0.00 $\\
and             & $100.0 \pm 0.00$ & $100.0 \pm 0.00$ & $100.0 \pm 0.00$ & --  & $100.0 \pm 0.00 $\\
relate\_left    & $100.0 \pm 0.00$ & $93.04 \pm 1.62$ & $100.0 \pm 0.00$ & $100.00 \pm 0.00$ & $100.0 \pm 0.00 $\\
filter\_large   & $100.0 \pm 0.00$ & $95.31 \pm 3.34$ & $100.0 \pm 0.00$ & --   & $100.0 \pm 0.00 $\\
query\_shape    & $100.0 \pm 0.00$ & $100.0 \pm 0.00$ & $100.0 \pm 0.00$ & $98.95 \pm 1.48$  & $100.0 \pm 0.00 $\\
same\_color     & $100.0 \pm 0.00$ & $100.0 \pm 0.00$ & $100.0 \pm 0.00$ & $99.84 \pm 0.48$ &$100.0 \pm 0.00 $ \\
\bottomrule
\end{tabular}
\label{tab:ablation_mending_CLEVR}
\end{table}

\begin{table}[tb]
\centering
\caption{\graph results without mending.}
\begin{tabular}{lrrrrr}
\toprule
$P$ & GPT-4o & DeepSeek & Mistral-Large & LLama3-70b  & Gemini-3\\
\midrule
cycle           & $86.28 \pm 2.83$ & $88.91 \pm 0.40$ & $83.86 \pm 2.75$ & --  & $90.18 \pm 1.20$\\
shortest\_path  & $89.00 \pm 0.81$ & $84.43 \pm 0.00$ & $89.40 \pm 0.00$ & --  & $91.53 \pm 0.00$\\
paths           & $86.04 \pm 1.21$ & $84.83 \pm 0.00$ & -- & --  & $92.92 \pm 0.00$\\
count\_nodes    & $84.43 \pm 0.00$ & $86.92 \pm 2.48$ & -- & --  & $90.13 \pm 0.00$\\
adjacent        & $89.40 \pm 0.00$ & $89.35 \pm 0.11$ & $89.35 \pm 0.11$ & $76.91 \pm 5.58$  & $99.33 \pm 0.00$   \\
adjacent\_to    & $89.40 \pm 0.00$ & $89.40 \pm 0.00$ & -- & $88.19 \pm 0.00$  & $95.33 \pm 0.53$\\
same\_line      & $89.35 \pm 0.11$ & $89.40 \pm 0.00$ & $86.93 \pm 1.92$ & -- & $97.21 \pm 1.42$\\
common\_station & $87.38 \pm 0.33$ & $88.51 \pm 0.70$ & -- & --  & $90.32 \pm 0.00$\\
line\_names     & $89.40 \pm 0.00$ & $89.40 \pm 0.00$ & -- & --  & $92.32 \pm 0.00$\\
line\_count     & $89.24 \pm 0.32$ & $89.24 \pm 0.32$ & -- & --  & $99.19 \pm 0.12$\\
exist           & $75.52 \pm 3.11$ & $89.40 \pm 0.00$ & -- & --  & $90.11 \pm 0.00$\\
\bottomrule
\end{tabular}
\label{tab:ablation_mending_CLEGR}

\end{table}

\begin{table}[tb]
\centering
\caption{GQA without Chain-of-Thought (CoT).}
\begin{tabular}{lrrrrr}
\toprule
$P$ & GPT-4o & DeepSeek & Mistral-Large & LLama3-70b & Gemini-3 \\
\midrule
query           & $96.54 \pm 2.98$ & $94.25 \pm 5.96$ & $95.34 \pm 0.51$    & $80.26 \pm 6.94$ & $98.92 \pm 0.00$ \\
exist           & $99.28 \pm 0.96$ & $99.98 \pm 0.00$ & $98.88 \pm 1.67$    & $87.39 \pm 2.44$ & $100.0 \pm 0.00$ \\
or              & $100.0 \pm 0.00$ & $100.0 \pm 0.00$ & $100.0 \pm 0.00$    & $100.0 \pm 0.00$ & $100.0 \pm 0.00$ \\
filter          & $97.24 \pm 3.38$ & $100.0 \pm 0.00$ & --                  & $89.27 \pm     8.61$ & $100.0 \pm 0.00$ \\
choose\_attr    & $99.02 \pm 2.06$ & $99.80 \pm 0.06$ & $99.80 \pm 0.06$    & $99.85 \pm 0.14$ & $ 99.83 \pm 0.00$ \\
verify\_rel     & $97.62 \pm 1.75$ & $100.0 \pm 0.01$ & $95.24 \pm 1.29$    & $97.41 \pm 2.34$  & $100.0 \pm 0.00$\\
select          & $96.06 \pm 3.41$ & $97.24 \pm 3.38$ & $81.12 \pm 35.79$   & -- &  $100.0 \pm 0.00$\\
negate          & $98.85 \pm 0.58$ & $99.71 \pm 0.58$ & $99.80 \pm 0.35$    & $98.56 \pm 0.00$ & $100.0 \pm 0.00$ \\
relate          & $95.38 \pm 8.69$ & $98.92 \pm 1.34$ & $91.10 \pm 11.73$   & $76.43 \pm 11.78$ & $99.94 \pm 0.12$\\
two\_different  & $100.0 \pm 0.00$ & $100.0 \pm 0.00$ & --                  & $99.88 \pm 0.20$ & $100.0 \pm 0.00$\\
two\_same       & $98.62 \pm 2.76$ & $100.0 \pm 0.00$ & --                  & $95.23 \pm 3.62$ &$100.0 \pm 0.00$ \\
\bottomrule
\end{tabular}
\label{tab:ablation_cot_GQA}
\end{table}

\begin{table}[tb]
\centering
\caption{CLEVR results without Chain-of-Thought (CoT).}
\begin{tabular}{lrrrrr}
\toprule
$P$ & GPT-4o & DeepSeek & Mistral-Large & LLama3-70b  & Gemini-3\\
\midrule
exist           & $99.72 \pm 0.56$ & $100.0 \pm 0.00$ & $96.72 \pm 0.56$ & $96.79 \pm 4.17$ &$100.0 \pm 0.00$ \\
unique          & $65.47 \pm 31.71$ & $100.0 \pm 0.00$ & -- & -- & $100.0 \pm 0.00$\\
count           & $100.0 \pm 0.00$ & $98.64 \pm 0.57$ & $98.41 \pm 0.00$ & -- &$100.0 \pm 0.00$ \\
equal\_integer  & $100.0 \pm 0.00$ & $100.0 \pm 0.00$ & $100.0 \pm 0.00$ & -- & $100.0 \pm 0.00$ \\
and             & $99.63 \pm 0.73$ & $100.0 \pm 0.00$ & $100.0 \pm 0.00$ & $100.0 \pm 0.00$  & $100.0 \pm 0.00$\\
relate\_left    & $100.0 \pm 0.00$ & $100.0 \pm 0.00$ & $100.0 \pm 0.00$  & $98.71 \pm 2.58$ & $100.0 \pm 0.00$\\
filter\_large   & $100.0 \pm 0.00$ & $100.0 \pm 0.00$ & $100.0 \pm 0.00$  & -- & $100.0 \pm 0.00$\\
query\_shape    & $100.0 \pm 0.00$ & $100.0 \pm 0.00$ & $100.0 \pm 0.00$  & $98.30 \pm 1.40$ &$100.0 \pm 0.00$ \\
same\_color     & $99.27 \pm 1.46$ & $100.0 \pm 0.00$ & $98.06 \pm 1.32$ & $100.0 \pm 0.00$ &$100.0 \pm 0.00$ \\
\bottomrule
\end{tabular}
\label{tab:ablation_cot_CLEVR}
\end{table}

\begin{table}[tb]
\centering
\caption{\graph results without Chain-of-Thought (CoT).}
\begin{tabular}{lrrrrr}
\toprule
$P$ & GPT-4o & DeepSeek & Mistral-Large & LLama3-70b & Gemini-3 \\
\midrule
cycle           & $87.60 \pm 2.41$ & $89.24 \pm 0.32$ & $59.40 \pm 22.62$ & -- &$ 99.19 \pm 0.00$\\
shortest\_path  & $84.51 \pm 6.80$ & $88.16 \pm 2.15$ & $83.90 \pm 9.53$ & -- & $100.0 \pm 0.00$\\
paths           & $80.22 \pm 8.18$ & $85.34 \pm 1.03$ & -- & -- & $ 95.62 \pm 4.20$\\
count\_nodes    & $81.02 \pm 6.82$ & $82.41 \pm 2.43$ & -- & -- & $100.0 \pm 0.00$\\
adjacent        & $89.40 \pm 0.00$ & $89.35 \pm 0.11$ & $89.02 \pm 0.26$ & $65.37 \pm 18.70$ & $100.0 \pm 0.00$\\
adjacent\_to    & $89.40 \pm 0.00$ & $89.40 \pm 0.00$ & -- & $88.19 \pm 0.00$ & $100.0 \pm 0.00$ \\
same\_line      & $86.53 \pm 5.61$ & $85.40 \pm 0.00$ & $84.22 \pm 0.32$ & -- & $100.0 \pm 0.00$\\
common\_station & $87.90 \pm 0.21$ & $88.64 \pm 0.16$ & -- & -- & $99.79 \pm 0.43$\\
line\_names     & $89.00 \pm 0.81$ & $89.40 \pm 0.00$ & --  & --  & $100.0 \pm 0.00$\\
line\_count     & $89.24 \pm 0.32$ & $80.00 \pm 9.00$ & --  & --  & $100.0 \pm 0.00$\\
exist           & $77.05 \pm 3.78$ & $89.40 \pm 0.00$ & --& -- &  $100.0 \pm 0.00$\\
\bottomrule
\end{tabular}
\label{tab:ablation_cot_CLEGR}
\end{table}

\begin{table}[tb]
\centering
\caption{GQA Performance without multiprompt.}
\begin{tabular}{lrrrrr}
\toprule
$P$ & GPT-4o & DeepSeek & Mistral-Large & LLama3-70b  & Gemini-3\\
\midrule
query           & $92.71 \pm 0.41$ & $98.92 \pm 0.00$ & $96.18 \pm 6.01$ & $70.06 \pm 22.05$ & $98.92 \pm 0.02$ \\
exist           & $99.52 \pm 0.95$ & $99.98 \pm 0.01$ & $100.0 \pm 0.00$ & $87.80 \pm 2.07$ & $100.0 \pm 0.00$\\
or              & $100.0 \pm 0.00$ & $100.0 \pm 0.00$ & $100.0 \pm 0.00$ & $98.29 \pm 1.98$ & $100.0 \pm 0.00$ \\
filter          & $95.27 \pm 6.45$ & $100.0 \pm 0.00$ & $100.0 \pm 0.00$ & $82.29 \pm 0.09$ & $100.0 \pm 0.00$\\
choose\_attr    & $99.83 \pm 0.00$ & $98.49 \pm 2.68$ & $93.82 \pm 1.21$ & $98.61 \pm 2.28$ & $99.87 \pm 0.07 $\\
verify\_rel     & $98.94 \pm 1.26$ & $99.84 \pm 0.31$ & $99.84 \pm 0.31$ & $96.18 \pm 2.34$ &$100.0 \pm 0.00$\\
select          & $81.68 \pm 36.0$& $100.0 \pm 0.00$ & $100.0 \pm 0.00$ & $51.35 \pm 41.7$ & $100.0 \pm 0.00$\\
negate          & $98.56 \pm 0.00$ & $95.71 \pm 0.28$ & $99.84 \pm 0.32$ & $98.56 \pm 0.00$ &$100.0 \pm 0.00$ \\
relate          & $98.74 \pm 2.30$ & $99.99 \pm 0.02$ & $96.43 \pm 2.91$ & $68.03 \pm 0.00$ & $100.0 \pm 0.00$ \\
two\_different  & $100.0 \pm 0.00$ & $100.0 \pm 0.00$ & $100.0 \pm 0.00$ & $99.89 \pm 0.19$ & $100.0 \pm 0.00$\\
two\_same       & $100.0 \pm 0.00$ & $100.0 \pm 0.00$ & $100.0 \pm 0.00$ & $99.68 \pm 0.27$ & $100.0 \pm 0.00$\\
\bottomrule
\end{tabular}
\label{tab:ablation_multiprompt_GQA}
\end{table}

\begin{table}[tb]
\centering
\caption{CLEVR results without multiprompt.}
\begin{tabular}{lrrrrr}
\toprule
$P$ & GPT-4o & DeepSeek & Mistral-Large & LLama3-70b & Gemini-3 \\
\midrule
exist           & $100.0 \pm 0.00$ & $99.75 \pm 0.51$ & $100.0 \pm 0.00$ & $87.98 \pm 3.98$ & $100.0 \pm 0.00$ \\
unique          & $71.68 \pm 34.68$& $43.04 \pm 27.68$ & -- &  $29.11 \pm 0.18$  & $100.0 \pm 0.00$\\
count           & $100.0 \pm 0.00$ & $99.44 \pm 0.70$ & -- & --  & $100.0 \pm 0.00$\\
equal\_integer  & $100.0 \pm 0.00$ & $100.0 \pm 0.00$ & $100.0 \pm 0.00$ & $93.89 \pm 5.03$ & $100.0 \pm 0.00$\\
and             & $100.0 \pm 0.00$ & $100.0 \pm 0.00$ & $100.0 \pm 0.00$ & $98.70 \pm 2.59$  & $100.0 \pm 0.00$\\
relate\_left    & $100.0 \pm 0.00$ & $100.0 \pm 0.00$ & $100.0 \pm 0.00$ & $100.00 \pm 0.00$  & $100.0 \pm 0.00$\\
filter\_large   & $100.0 \pm 0.00$ & $100.0 \pm 0.00$ & $100.0 \pm 0.00$ & $87.42 \pm 15.41$  & $100.0 \pm 0.00$\\
query\_shape    & $100.0 \pm 0.00$ & $100.0 \pm 0.00$ & $99.75 \pm 0.51$ & $95.25 \pm 6.60$  & $100.0 \pm 0.00$\\
same\_color     & $100.0 \pm 0.00$ & $100.0 \pm 0.00$ & $99.27 \pm 1.46$ & $100.00 \pm 0.00$  & $100.0 \pm 0.00$\\
\bottomrule
\end{tabular}
\label{tab:ablation_multiprompt_CLEVR}
\end{table}

\begin{table}[tb]
\centering
\caption{\graph results without multiprompt.}
\begin{tabular}{lrrrrr}
\toprule
$P$ & GPT-4o & DeepSeek & Mistral-Large & LLama3-70b & Gemini-3 \\
\midrule
cycle           & $88.75 \pm 0.32$  & $88.97 \pm 0.37$  & --               & -- & $99.19 \pm 0.00$ \\
shortest\_path  & $86.92 \pm 2.48$  & $84.43 \pm 0.00$  & $88.41 \pm 1.99$ & --  & $100.0 \pm 0.00$\\
paths           & --                & --                & $82.15 \pm 6.72$ & --  &  $84.12 \pm 10.5$\\
count\_nodes    & --                & --                & $86.11 \pm 1.68$ & --  & $100.0 \pm 0.00$\\
adjacent        & --                & --                & --               & $74.82 \pm 19.26$  & $99.60 \pm 0.33$\\
adjacent\_to    & --                & $88.67 \pm 0.59$  & $88.92 \pm 0.59$ & --  & $100.0 \pm 0.00$\\
same\_line      & --                & --                & $88.56 \pm 0.97$ & $80.08 \pm 6.64$  & $100.0 \pm 0.00$\\
common\_station & $88.08 \pm 0.60$  & $87.11 \pm 1.20$  & $86.57 \pm 1.74$ & $86.98 \pm 2.13$  & $99.87 \pm 0.27$\\
line\_names     & --                & --                & $87.36 \pm 4.08$ & --  &  $100.0 \pm 0.00$\\
line\_count     & --                & $89.08 \pm 0.40$  & --               & --  & $100.0 \pm 0.00$\\
exist           & --                & --                & $87.87 \pm 3.06$ & --  & $100.0 \pm 0.00$\\
\bottomrule
\end{tabular}
\label{tab:ablation_multiprompt_CLEGR}
\end{table}

\section{Composite Rules for GQA}\label{apdx:composite}
This section lists the ASP definitions for all synthesized predicates that were not part of the initial theory, described in Section~\ref{ssection:composite}.
These predicates emerged during distillation and were successfully validated against test data, demonstrating compositional reasoning capabilities.

\begin{itemize}
    \item For \texttt{connected}, we select two objects from the scene and ensure the presence of a path between them through one or more intermediate objects, \iec we synthesize relational chains by adding \texttt{has\_rel(A, connects, B)} relations (\egc object A is next to object B, and B is to the left of C) to form paths of length greater than two. The query then tests whether such indirect connections exist.
    \begin{verbatim}
bool(T, yes) :- connected(T, T0, T1), state(T0, ID0), 
                state(T1, ID1), path(ID0, ID1).
bool(T, no)  :- connected(T, T0, T1), not bool(T, yes).
path(X,Y)    :- has_rel(X, connects, Y).
path(X,Y)    :- has_rel(X, connects, Z), path(Z,Y).
    \end{verbatim}
    \item For \texttt{isolated}, we identify an object and systematically remove all of its relationships from the scene graph, making it structurally disconnected. The associated query asks whether the object is isolated.
    \begin{verbatim}
bool(TO, yes) :- isolated(TO, TI), state(TI, ID), 
                 not has_rel(ID, _, _), not has_rel(_, _, ID).
bool(TO, no)  :- isolated(TO, TI), not bool(TO, yes).
        \end{verbatim}
    \item For \texttt{count\_class}, we leave the scene unchanged and simply construct a query that counts the number of objects belonging to a given class. We note that counting is not part of the original GQA questions. 
    \begin{verbatim}
count(TO, N) :- count_class(TO, TI, CLASS), 
                N = #count { ID : has_attr(ID, class, CLASS) }.
ans(V)       :- end(TO), count(TO,V).
    \end{verbatim}
\end{itemize}

\section{Pruning Heuristic Tables}\label{apdx:pruning}
This final appendix section provides pruning-related statistics for each model and predicate, which are visualized in Section~\ref{sec:redundancy}. 
Tables~\ref{tab:pruning_GQA},~\ref{tab:pruning_CLEVR} and~\ref{tab:pruning_CLEGR} show the results for GQA, CLEVR and \graph respectively.
We report how many rules were pruned during distillation, with the minimum and maximum between parenthesis. 
This helps to assess over-generation tendencies and possible can be used for feedback-loop improvements.

\begin{table}[tb]
\centering
\caption{Number of rules pruned on GQA.}
\begin{tabular}{lrrrrr}
\toprule
$P$ & GPT-4o & DeepSeek & Mistral-Large & LLama3-70b  & Gemini-3 \\
\midrule
query           & 1.2 (0, 4)  & 3.0 (0, 6)  & 0.0 (0, 0) & 2.0 (0, 7)  & 0.0 (0, 0) \\
exist           & 0.8 (0, 2)  & 3.2 (0, 9)  & 0.25 (0, 1)& 0.25 (0, 1) & 0.0 (0, 0) \\
or              & 1.0 (1, 1)  & 2.2 (1, 5)  & 1.0 (1, 1) & 1.0 (0, 2)  & 1.0 (1, 1) \\
filter          & 7.0 (0, 15) & 3.0 (0, 9)  & 0.0 (0, 0) & 1.25 (0, 5) & 0.0 (0, 0) \\
choose\_attr    & 6.8 (3, 21) & 1.75 (0, 5) & 1.6 (0, 2) & 1.0 (0, 3)  & 0.0 (0, 0) \\
verify\_rel     & 3.2 (2, 4)  & 4.5 (1, 8)  & 0.8 (0, 2) & 0.0 (0, 0)  & 1.0 (1, 1) \\
select          & 0.8 (0, 2)  & 1.4 (0, 6)  & 0.5 (0, 1) & 0.5 (0, 2)  & 0.0 (0, 0) \\
negate          & 0.6 (0, 1)  & 1.0 (1, 1)  & 0.6 (0, 3) & 0.25 (0, 1) & 0.0 (0, 0) \\
relate          & 4.2 (1, 11) & 4.2 (0, 12) & 0.0 (0, 0) & 1.33 (0, 4) & 0.0 (0, 0) \\
two\_different  & 0.0 (0, 0)  & 2.2 (0, 8)  & 0.0 (0, 0) & 0.0 (0, 0)  & 0.0 (0, 0) \\
two\_same       & 0.2 (0, 1)  & 2.5 (0, 10) & 0.0 (0, 0) & 0.0 (0, 0)  & 0.0 (0, 0) \\
\bottomrule
\end{tabular}
\label{tab:pruning_GQA}
\end{table}

\begin{table}[tb]
\centering
\caption{Number of rules pruned on CLEVR.}
\begin{tabular}{lrrrrr}
\toprule
$P$ & GPT-4o & DeepSeek & Mistral-Large & LLama3-70b & Gemini-3 \\
\midrule
exist           & 5.8 (4, 8) & 3.0 (2, 4) & 6.0 (4, 10) & 0.5 (0, 3) & 0.0 (0, 0) \\
unique          & 8.8 (5, 15) & 2.4 (1, 6) & -- & --  & 0.0 (0, 0) \\
count           & 7.4 (4, 12) & 2.0 (0, 6) & 2.2 (0, 8) & -- & 0.0 (0, 0) \\
equal\_integer  & 6.2 (2, 12) & 5.2 (3, 8) & 9.8 (8, 12) & --  & 0.0 (0, 0) \\
and             & 6.4 (4, 10) & 1.0 (1, 1) & 1.2 (0, 4) &  0.42 (0, 1) & 0.0 (0, 0) \\
relate\_left    & 5.0 (1, 10) & 1.0 (1, 1) & 6.4 (4, 8) & 1.0 (0, 4)  & 0.0 (0, 0) \\
filter\_large   & 6.8 (5, 10) & 2.4 (1, 4) & 6.6 (3, 9) & --  & 0.0 (0, 0) \\
query\_shape    & 2.8 (1, 5) & 1.2 (1, 2) & 5.4 (3, 8) &  0.14 (0, 1) & 0.0 (0, 0) \\
same\_color      & 2.8 (1, 8) & 1.0 (1, 1) & 4.4 (1, 8) &  1.6 (0, 4) & 0.0 (0, 0) \\
\bottomrule
\end{tabular}
\label{tab:pruning_CLEVR}
\end{table}

\begin{table}[tb]
\centering
\caption{Number of rules pruned on \graph.}
\begin{tabular}{lrrrrr}
\toprule
$P$ & GPT-4o & DeepSeek & Mistral-Large & LLama3-70b & Gemini-3 \\
\midrule
cycle           & 1.8 (1, 3) & 4.0 (2, 8) & 2.4 (2, 4) & -- & 0.8 (0, 1) \\
shortest\_path  & 3.0 (0, 4) & 3.0 (2, 4) & 1.0 (0, 4) & -- & 0.4 (0, 1) \\
paths           & 4.0 (2, 6) & 4.0 (3, 5) & --         & -- & 1.0 (1, 1) \\
count\_nodes     & 2.8 (0, 12)& 2.0 (2, 2) & 0.0 (0, 0) & -- & 0.0 (0, 0) \\
adjacent        & 2.6 (2, 3) & 2.6 (2, 4) & 3.8 (1, 7) & 0.75 (0, 3) & 0.2 (0, 1) \\
adjacent\_to     & 2.4 (2, 3) & 3.0 (2, 4) & 0.0 (0, 0) & 0.75 (0, 3) & 0.2 (0, 1) \\
same\_line       & 3.0 (2, 6) & 1.0 (1, 1) & 2.6 (2, 3) & -- & 0.5 (0, 1) \\
common\_station  & 2.8 (2, 4) & 7.2 (2, 14)& 2.0 (0, 5) & -- & 0.8 (0, 1) \\
line\_names      & 2.0 (2, 2) & 1.0 (1, 1) & 0.11 (0, 1)& -- & 0.0 (0, 0) \\
line\_count      & 2.0 (2, 2) & 1.0 (1, 1) & 0.0 (0, 0) & -- & 0.0 (0, 0) \\
exist           & 2.2 (0, 4) & 1.0 (1, 1) & 0.0 (0, 0) & -- & 0.2 (0, 1) \\
\bottomrule
\end{tabular}
\label{tab:pruning_CLEGR}
\end{table}

\end{document}